\documentclass[11pt]{article}

\usepackage[final]{acl}

\usepackage{times}
\usepackage{latexsym}

\usepackage[T1]{fontenc}

\usepackage[utf8]{inputenc}

\usepackage{microtype}

\usepackage{inconsolata}

\usepackage{graphicx}
\usepackage{hyperref}
\usepackage{url}
\usepackage{microtype}
\usepackage{graphicx}
\usepackage{booktabs}
\usepackage{enumitem}
\usepackage{hyperref}
\usepackage{wrapfig}
\usepackage{algorithm}
\usepackage{algpseudocode}  
\usepackage{marvosym}
\usepackage{subcaption}
\usepackage{tikz}
\usepackage{pgfplots}
\pgfplotsset{compat=1.18}
\usepackage{adjustbox}
\usetikzlibrary{positioning,calc,arrows.meta,fit,shapes.misc,decorations.pathreplacing}
\usepackage{hyperref}
\usepackage{url}
\usepackage{graphicx}
\usepackage[table]{xcolor}  
\usepackage{booktabs}  
\usepackage{multirow}   
\usepackage{graphicx}    
\usepackage{array}       
\usepackage{adjustbox}
\usepackage{subcaption}
\usepackage{xcolor}
\usepackage{amsmath}
\usepackage{amssymb}
\usepackage[most]{tcolorbox}
\usepackage[frozencache=true,cachedir=minted-cache]{minted}
\tcbuselibrary{minted}

\newcolumntype{P}[1]{>{\centering\arraybackslash}m{#1}}

\newcommand{\nocontentsline}[3]{}
\newcommand{\tocless}[2]{\bgroup\let\addcontentsline=\nocontentsline#1{#2}\egroup}

\newcommand{\mytcbmanual}[3]{%
\begin{tcolorbox}[
  enhanced,
  breakable,
  colback=white,
  colframe=beige,
  title=#1,
  fonttitle=\bfseries,
  listing only,
  listing engine=minted,
  minted language=#3,
  minted options={
    fontsize=\scriptsize,
    breaklines,
    autogobble,
    tabsize=2
  }
]
#2
\end{tcolorbox}
}

\usepackage{multirow, booktabs, pifont, xcolor}

\definecolor{beige}{RGB}{204,182,146}
\definecolor{lightbeige}{RGB}{254,242,206}
\definecolor{steelblue}{RGB}{78,111,136}

\title{Agent-GWO: Collaborative Agents for Dynamic Prompt Optimization in Large Language Models}

\author{
  \textbf{Xudong Wang}\textsuperscript{1},\ 
  \textbf{Chaoning Zhang}\textsuperscript{2$^{*}$},\ 
  \textbf{Chenghao Li}\textsuperscript{2},\ 
  \textbf{Shuxu Chen}\textsuperscript{1},\\
  \textbf{Qigan Sun}\textsuperscript{1},\ 
  \textbf{Jiaquan Zhang}\textsuperscript{2},\ 
  \textbf{Fachrina Dewi Puspitasari}\textsuperscript{2},\ 
  \textbf{Tae-Ho Kim}\textsuperscript{3},\\
  \textbf{Jiwei Wei}\textsuperscript{2},\ 
  \textbf{Malu Zhang}\textsuperscript{2},\  
  \textbf{Guoqing Wang}\textsuperscript{2},\ 
  \textbf{Yang Yang}\textsuperscript{2},\
  and \textbf{Heng Tao Shen}\textsuperscript{4}\\
  \\
  \textsuperscript{1}Kyung Hee University\quad
  \textsuperscript{2}University of Electronic Science and Technology of China\quad
  \\
  \textsuperscript{3}Nota Inc.\quad
  \textsuperscript{4} Tongji University \quad
  \\
  \texttt{wangcurry33@gmail.com \Letter chaoningzhang1990@gmail.com}
}

\begin{document}
\maketitle

\begin{abstract}
Large Language Models (LLMs) have demonstrated strong capabilities in complex reasoning tasks, while recent prompting strategies such as Chain-of-Thought (CoT) have further elevated their performance in handling complex logical problems. Despite these advances, high-quality reasoning remains heavily reliant on manual static prompts and is sensitive to decoding configurations and task distributions, leading to performance fluctuations and limited transferability. Existing automatic prompt optimization methods typically adopt single-agent local search, failing to simultaneously optimize prompts and decoding hyperparameters within a unified framework to achieve stable global improvements. To address this limitation, we propose Agent-GWO, a dynamic prompt optimization framework for complex reasoning. Specifically, we unify prompt templates and decoding hyperparameters as inheritable agent configurations. By leveraging the leader-follower mechanism of the Grey Wolf Optimizer (GWO), we automatically select three leader agents ($\alpha$, $\beta$, and $\delta$) to guide the collaborative updates of the remaining agents, enabling iterative convergence toward robust optimal reasoning configurations that can be seamlessly integrated for inference. Extensive experiments on multiple mathematical and hybrid reasoning benchmarks across diverse LLM backbones show that Agent-GWO consistently improves accuracy and stability over existing prompt optimization methods. \textit{The code will be released publicly.}
\end{abstract}

\section{Introduction}
LLMs have recently achieved substantial progress across a range of reasoning tasks, particularly when leveraged with prompting strategies such as Chain-of-Thought (CoT) that elicit step-by-step reasoning ~\cite{wei2022chain,kojima2022large}. Despite these gains, strong average performance does not necessarily translate into reliable reasoning behavior. As problems become longer and more complex, accuracy often degrades significantly ~\cite{cao2024worst,li2025patterns}. More critically, performance remains highly sensitive to prompt wording, exemplar ordering, and contextual perturbations: minor edits or re-orderings that leave the underlying task unchanged can nonetheless precipitate large swings in performance ~\cite{chatterjee2024posix,ariyani2025there,madani2026inclusion}. This prompt's fragility makes it difficult to distinguish genuinely robust reasoning steps from those that are merely accidentally correct. Furthermore, early deviations in the reasoning chain can cascade through subsequent steps, severely compromising the reliability of multi-step solutions ~\cite{yue2023large,he2025can,wang2025think}.

\begin{figure}[t!]
    \centering
    \includegraphics[width=1\linewidth]{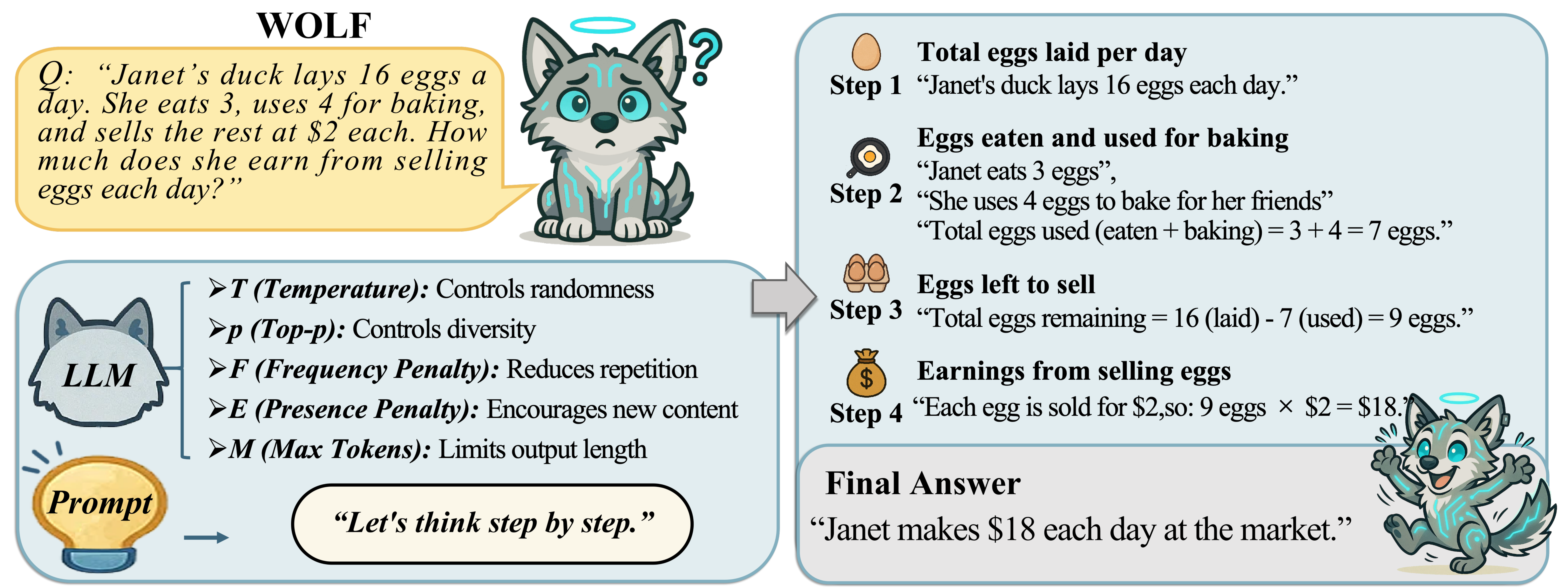}
    \caption{
        GWO abstracts each LLM agent as a ``wolf'' which is characterized by hyperparameters \& reasoning prompt.
        These two guide the behavior of the LLMs during the optimization process.
    }
    \label{fig:agentdef}
\end{figure}

To address the above reliability challenge, prior work has proposed a range of inference-time enhancement strategies that improve the robustness of single-pass decoding primarily through trajectory generation and search, such as self-consistency sampling, multi-path exploration, and tree/graph-structured reasoning~\cite{madaan2023self,zhang2024aflow,teng2025atom,zhang2026tda,zhang2026learning,huang2026semanticspaceexplorationexploitationrlvr}. Although these methods often deliver gains without additional training, they are typically built upon manually crafted static prompts. As a result, performance can still fluctuate markedly under task transfer, contextual perturbations, or minor prompt rewrites, leaving prompt brittleness and poor reusability largely unresolved\cite{jin2024impact}. Moreover, repeated sampling and structured search commonly introduce extra inference cost and system complexity, which limit their applicability in resource-constrained or low-latency settings~\cite{yue2023large,he2025can}. In contrast, training-based adaptation can offer more systematic improvements, but usually requires extra data, training iterations, and engineering effort, making it less suitable as a lightweight, general-purpose inference-stage solution~\cite{lester2021power,hu2022lora}.
Overall, existing work largely improves reasoning trajectories under fixed prompts, rather than providing a reproducible and scalable mechanism that can automatically optimize both the prompting strategy and the decoding behavior at inference time.

Motivated by this gap, we ask a key question: \textbf{\textit{at inference time, can we reduce reliance on manual prompt trial-and-error and automatically discover prompt-configuration pairs that are more stable and better aligned with task requirements?}} Our core observation is that, for a given frozen model, reasoning quality is mainly governed by two controllable factors: the prompt template (how the reasoning process is organized and constrained) and the decoding configuration (e.g., stochasticity, repetition penalties, and length budgets). Systematically exploring this joint space via population-based search offers the potential to uncover reasoning strategies that are both more stable and more task-adaptive.

To this end, we propose Agent-GWO, a dynamic prompt optimization framework for complex reasoning. As illustrated in Figure~\ref{fig:agentdef}, each agent is defined by a reasoning prompt template together with a decoding configuration, and a population of such agents is optimized collaboratively in an iterative manner. At each iteration, multiple agents generate reasoning traces in parallel and are ranked according to task-level performance. We then design a hierarchical collaborative update mechanism that enables structured information sharing among agents. By exploiting leader--follower dynamics inspired by the Grey Wolf Optimizer, this mechanism allows top-performing agents to guide population-level updates and steer the evolution of the remaining agents. This design preserves population diversity to encourage exploration while ensuring stable and controllable convergence toward high-quality reasoning configurations. The resulting optimized configuration can be directly reused at inference time without any additional training.

Our contributions are summarized as follows:
\begin{itemize}
\item We propose an automated inference-time adaptation paradigm that jointly optimizes prompt templates and decoding parameters at test time, improving both accuracy and stability without additional training.
\item We design a hierarchical leader--follower swarm mechanism inspired by GWO to stabilize multi-agent prompt search and reduce sensitivity to prompt variants.
\item Extensive experiments on mathematical and hybrid reasoning benchmarks across diverse LLM backbones show consistent gains under controllable inference budgets over strong manual baselines and prior prompt optimization methods.
\end{itemize}

\section{Related Work}

\subsection{Prompt Engineering and Static Chain-of-Thought Methods}
Prompting is a widely used approach for adapting LLMs in zero- and few-shot settings~\cite{brown2020language}. Chain-of-Thought (CoT) prompting~\cite{wei2022chain} elicits step-by-step reasoning and has inspired extensions such as self-consistency~\cite{wang2022self}, least-to-most decomposition~\cite{zhou2022least}, and structured variants including Self-Ask~\cite{press2022selfask}, Program-of-Thoughts~\cite{chen2022pot}, and Graph-of-Thoughts~\cite{liu2023got}.

Despite their success, these methods still rely heavily on task-specific prompt design~\cite{zhang2022automatic} and can be sensitive to prompt phrasing and exemplar ordering~\cite{lu2022fantastically}. Moreover, long reasoning traces and multi-sample strategies often increase inference cost~\cite{zhou2022least,zheng2025joint,cao2026language,zheng2026llava,zhang2026text}. Recent work therefore studies automatic prompt optimization to reduce manual trial-and-error and improve transferability, e.g., via evolutionary or gradient-free search and feedback-driven refinement. Our work follows this direction but focuses on population-based, leader-guided optimization over both prompt templates and decoding configurations under frozen model weights.

\subsection{Collaborative Reasoning with Multi-Agent LLMs}
Multi-Agent Systems (MAS) provide a natural framework for collaborative reasoning through coordination and role specialization~\cite{wooldridge1995intelligent,ferber1999multi}. With LLMs, MAS has been used for planning and reasoning via agent communication and iterative interaction~\cite{naveed2023comprehensive,kumar2024large,xi2025rise}, with representative frameworks such as AutoGen~\cite{wu2023autogen} and Chain of Agents~\cite{zhang2024chain}. Prior work explores modular agent designs~\cite{liu2025advances,zhang2026lightweightllmagentmemory}, debate-style aggregation~\cite{du2023improving,xu2026rcbsfmultiagent}, and communication/coordination protocols~\cite{fioretto2018distributed,fipa2002fipa,jiang2018learning,yang2026evotool,yang2026tooltree}, and applies them to diverse domains~\cite{park2023generative,zhang2024generative,lim2024large,li2023econagent,zheng2026towards,wang2026streammecolongtermagentmemory}.

Complementary to this line, we cast multi-agent collaboration as an explicit iterative optimization process, where strong agents guide the update of others to improve task-specific prompting and decoding behaviors.

\begin{figure*}[ht!]
    \centering
    \includegraphics[width=\linewidth]{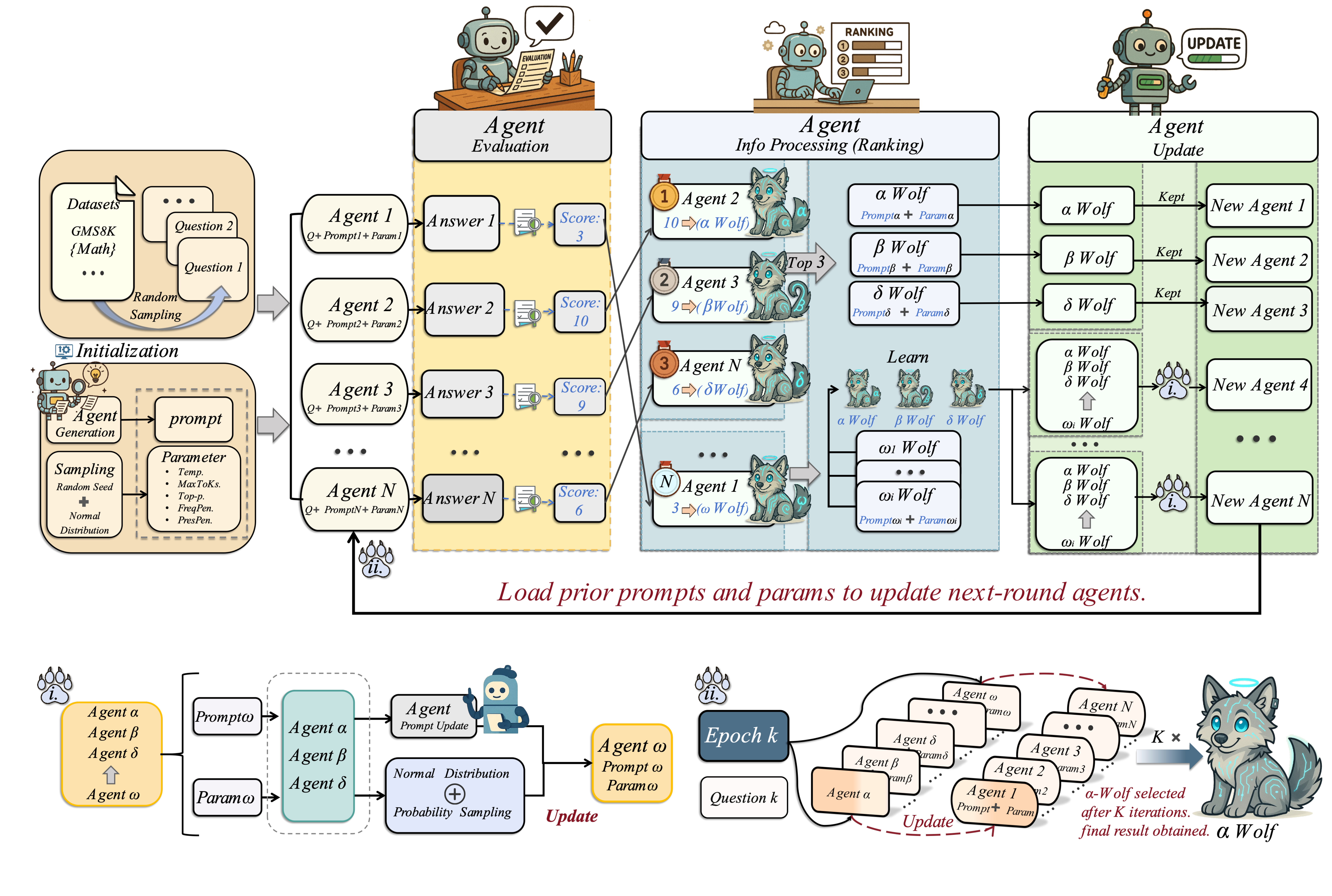}
    \caption{The overall Agent-GWO framework operates by having each agent process a dataset, Prompt, and parameters to produce outputs, which are then scored. Only the top-3 agents ($\alpha$, $\beta$, and $\delta$) are differentiated, while the rest (i.e., $\omega$) are updated using the prompt and parameters from the previous round. This iterative ranking and update process continues until termination, after which the final agent is set as the top-ranked $\alpha$ from the last iteration. \textbf{Note:} The left-bottom subfigure illustrates the learning process of each agent at each iteration and the right-bottom subfigure illustrates the optimization process.}
    \label{fig:enterlabel}
\end{figure*}

\section{Method}

\subsection{Theoretical Background and Problem Formulation}

\textbf{Grey Wolf Optimizer (GWO).}
GWO~\cite{mirjalili2014grey} is a population-based metaheuristic that maintains $N$ candidate solutions (``wolves'') and updates them using a leader--follower hierarchy.
At each iteration, the top three solutions are treated as leaders ($\alpha$, $\beta$, $\delta$), and other wolves ($\omega$) are guided toward them.
For a wolf $\mathbf{X}_i$ in a $D$-dimensional space, the standard GWO update averages the three leader-guided positions:
\begin{equation}
    \mathbf{X}_i(t+1) = \tfrac{\mathbf{X}_i^{(\alpha)}(t+1)+\mathbf{X}_i^{(\beta)}(t+1)+\mathbf{X}_i^{(\delta)}(t+1)}{3}
\end{equation}
where $\mathbf{X}_i^{j}(t{+}1)$ denotes the leader-guided update induced by leader $j\in\{\alpha,\beta,\delta\}$ using the encircling rule (Appendix~\ref{sec:gwo_details}).

\noindent \textbf{Agent Structure and Definition}. Suppose there are \(n\) agents. The \(j\)-th agent, denoted as \(\text{Agent}_j\), consists of a large language model \(\text{LLM}_j\) and a prompt template \(\text{prompt}_j\), i.e., \(\text{Agent}_j = \{\text{LLM}_j, \text{prompt}_j\}\). The language model \(\text{LLM}_j\) contains a shared model parameter set \(\boldsymbol{\theta}\) and an agent-specific hyperparameter set \(\boldsymbol{\eta}_j = \{T_j, p_j, F_j, E_j, M_j\}\), which represent temperature, top-\(p\) threshold, frequency penalty, presence penalty, and maximum token length, respectively. Therefore, \(\text{LLM}_j = \{\boldsymbol{\eta}_j, \boldsymbol{\theta}\}\), and \(\text{Agent}_j = \left\{ \left\{ \boldsymbol{\eta}_j, \boldsymbol{\theta} \right\}, \text{prompt}_j \right\}\).

\noindent \textbf{Hyperparameter Sampling.} To ensure diversity and stability in agent behavior, we design a hyperparameter sampling strategy for the set 
\(\boldsymbol{\eta}_j = \{t_j, p_j, f_j, e_j, m_j\}\), where a clipping function 
\(\text{clip}(x, [a, b]) = \max(a, \min(x, b))\) constrains sampled values within valid ranges, thereby maintaining controlled and consistent generation. Specifically, the temperature \(t_j\) and the top-\(p\) threshold \(p_j\) are independently drawn from normal distributions 
\(\mathcal{N}(\mu_t, \sigma_t^2)\) and \(\mathcal{N}(\mu_p, \sigma_p^2)\), then clipped to intervals \([a_t, b_t]\) and \([a_p, b_p]\), respectively. These parameters regulate the generation distribution: higher values enhance creativity and diversity, while lower values promote predictability and stability. The frequency penalty \(f_j\), sampled from \(\mathcal{N}(\mu_f, \sigma_f^2)\) and clipped to \([a_f, b_f]\), mitigates repetitiveness by penalizing tokens proportionally to their prior frequency. The presence penalty \(e_j\), sampled from \(\mathcal{N}(\mu_e, \sigma_e^2)\) and clipped to \([a_e, b_e]\), encourages novelty by uniformly penalizing any previously generated token, thus fostering the introduction of new words or concepts. Finally, the maximum token length \(m_j\) is set to a fixed constant \(c_m\) for tasks requiring predetermined length, while for adaptive-length tasks, it is uniformly drawn from a discrete set \(\mathcal{M}\), i.e., \(m_j \sim \text{Uniform}(\mathcal{M})\).

This sampling approach uses normal or uniform distributions to introduce controlled variability, while clipping ensures hyperparameters remain within reasonable bounds, optimizing creativity, stability, and content quality.

\noindent \textbf{Problem Formulation.}
We denote the reasoning task dataset as \(\mathcal{D} = \{q_1, q_2, \dots, q_N\}\), where each element \(q_i\) represents a specific question. The dataset consists of \(N\) question samples in total. During the reasoning process, given any question \(q \in \mathcal{D}\), a specific agent, denoted as \(\text{Agent}_j\), processes the question and outputs two components: first, a detailed chain of thought (denoted as \(\text{CoT}_j\)), which illustrates the step-by-step reasoning process taken by the agent; and second, a final answer (denoted as \(\text{Answer}_j\)) that the agent derives based on the chain of thought. This reasoning process can be formally expressed as a function:
\begin{equation}
    f(\text{Agent}_j, q) = (\text{CoT}_j, \text{Answer}_j),\quad q \in \mathcal{D}
\end{equation}

\paragraph{Evaluation Function.}
For verifiable reasoning benchmarks with gold labels, we define the fitness of agent $j$ as the exact-match accuracy on a validation batch $D_{\mathrm{val}}$:
\begin{equation}
\mathrm{FITNESS}_j
= \frac{1}{\lvert D_{\mathrm{val}} \rvert}
\sum_{q \in D_{\mathrm{val}}}
\mathbb{1}\!\left[\mathrm{Answer}_j(q) = y(q)\right].
\end{equation}

To stabilize leader selection, we compute auxiliary scores
$\bigl(s^{\mathrm{logic}},\, s^{\mathrm{creativity}},\, s^{\mathrm{complete}}\bigr)$
using a fixed LLM-judge prompt (Appendix~\ref{sec:LLM-judge}). These auxiliary scores are not used to replace accuracy, but only to rank agents when accuracies are tied. For non-verifiable tasks, we use the same judge to define
\begin{equation}
\mathrm{FITNESS}_j
= w_1 s^{\mathrm{logic}} + w_2 s^{\mathrm{creativity}} + w_3 s^{\mathrm{complete}},
\end{equation}
where the weights are calibrated on 200 human-labeled samples.
We set the calibrated weights to $(w_1=0.5,\, w_2=0.2,\, w_3=0.3)$, reflecting the higher importance of logical correctness. All judge-based scores are averaged over three independent judge seeds.

\subsection{Multi-Agent Grey Wolf Optimizer Framework}
As illustrated in Figure~\ref{fig:enterlabel}, we adapt the GWO to optimize LLM agent configurations by modeling each agent as a ``wolf'' in a population. Each agent $\mathrm{Agent}_j=(\boldsymbol{\eta}_j,\mathrm{prompt}_j)$ corresponds to a candidate solution, defined by its decoding hyperparameters and prompt template. The overall procedure follows GWO's leader--follower hierarchy, where the top-performing agents ($\alpha$, $\beta$, $\delta$) guide the updates of the remaining agents over multiple iterations. The full pseudocode is provided in Appendix~\ref{Pseudocode}.

\noindent\textbf{Initialization.} The algorithm initializes a population \(\mathcal{A}\) of \text{n} agents. For each agent \(\text{Agent}_j\), hyperparameters \(\boldsymbol{\eta}_j\) are randomly sampled from predefined ranges (e.g., temperature \(\in [0, 1]\), top-p \(\in [0, 1]\)) using a normal distribution. Similarly, the prompt \(\text{prompt}_j\) is sampled from a set of predefined CoT prompt templates or generated via a template generation function. This diverse initialization ensures broad exploration of the search space.

\noindent\textbf{Optimization Loop.} The optimization phase iterates \(K\) times, with each iteration evaluating and updating the agent population. For each agent \(\textit{Agent}_j\), a question \(q\) is sampled from \(\mathcal{D}\), and the LLM generates a CoT trace and answer using the agent's configuration. The fitness of the answer is computed using a carefully designed evaluation function, which assesses the response based on three crucial dimensions: the logical consistency of the reasoning process, the ingenuity of the reasoning approach, and the comprehensiveness of the reasoning content. This multi-dimensional evaluation ensures a thorough and accurate assessment of the answer's quality. The top three agents, denoted \(\alpha\), \(\beta\), and \(\delta\), are selected based on their fitness scores, representing the best, second-best, and third-best solutions, respectively.

Non-elite agents (\(\mathcal{A} \setminus \{\alpha, \beta, \delta\}\)) are updated to converge toward the top performers. For each hyperparameter \(\eta_j^{(k)}\) in agent \(\textit{Agent}_j\), new values are sampled from normal distributions centered at the top agents:
$\ X_r \sim \mathcal{N}(\eta_r^{(k)}, \sigma^2), \quad \text{for } r \in \{\alpha, \beta, \delta\},$
and combined as a weighted average:
$\eta_j^{(k)} = w_{\alpha} X_{\alpha} + w_{\beta} X_{\beta} + w_{\delta} X_{\delta},$ with weights satisfying \(w_{\alpha} > w_{\beta} > w_{\delta}\) and \(w_{\alpha} + w_{\beta} + w_{\delta} = 1\). This prioritizes the influence of the best-performing agent (\(\alpha\)) while maintaining diversity. The prompt is adapted by a structured LLM-driven update function \(\mathrm{PromptAdaptation}(\cdot)\). Formally, \(P_j^{(k+1)}=\mathrm{PromptAdaptation}(P_j^{(k)}, P_{\alpha}, P_{\beta}, P_{\delta}, \mathcal{I})\), where \(P_j^{(k)}\) denotes the current prompt of agent \(j\), \(P_{\alpha}, P_{\beta}, P_{\delta}\) are the prompts of the top-three elite agents, and \(\mathcal{I}\) is a fixed system instruction. Under \(\mathcal{I}\), the LLM performs lightweight edits on \(P_j^{(k)}\) while preserving the task objective and evaluation criteria. The allowed edits are limited to step re-ordering, semantically equivalent paraphrasing, and formatting or constraint adjustments.

\noindent\textbf{Overall procedure.}
Each iteration consists of generation, evaluation, ranking, and update, repeated for $K$ rounds until returning the final leader configuration.

\begin{table*}[ht!]
    \centering
    \renewcommand{\arraystretch}{1.1}
    \resizebox{\textwidth}{!}{
        \begin{tabular}{l|c|c|c|c|c|c|c|c|c|c|c|c}
            \toprule
            
            \multirow{2}{*}{\textbf{Model}} & 
            \multicolumn{6}{c}{\textbf{Math Reasoning Tasks}} & 
            \multicolumn{5}{c}{\textbf{Hybrid Reasoning Tasks}} &
            \multirow{2}{*}{\textbf{Avg}} \\
            \cmidrule(lr){2-7} \cmidrule(lr){8-12}
            
             & \textbf{GSM8K} & \textbf{MATH} & \textbf{SVAMP} & \textbf{MultiArith} & \textbf{ASDiv} & \textbf{AQUA} 
             & \textbf{MMLU} & \textbf{BBH} & \textbf{Date} & \textbf{CLUTRR} & \textbf{MATH\_MIX} \\
            \midrule

            \rowcolor{lightbeige} 
            \multicolumn{13}{c}{\textbf{GPT-4o-mini}} \\ 
            \midrule 
            CoT & 85.4\% & 74.8\% & 84.7\% & 89.5\% & 92.3\% & 65.3\% & 63.4\% & 66.6\% & 52.1\% & 66.2\% & 80.1\% & 74.6\% \\
            CoT-SC/n=5 & 89.9\% & 76.3\% & 85.8\% & 89.7\% & 93.3\% & 70.7\% & 67.1\% & 69.0\% & 54.7\% & 72.2\% & 81.6\% & 77.3\% \\
            Self-Refine & 88.9\% & 75.1\% & 85.0\% & 89.9\% & 92.2\% & 67.8\% & 64.8\% & 67.5\% & 53.0\% & 68.4\% & 83.8\% & 76.0\% \\
            AFlow & 93.5\% & 78.9\% & 88.7\% & 91.9\% & 93.4\% & 72.0\% & 69.5\% & 68.0\% & 66.0\% & 72.8\% & 85.3\% & 80.0\% \\
            ToT & 94.9\% & 77.8\% & 89.6\% & 89.8\% & 93.6\% & 72.4\% & 71.6\% & 69.9\% & 69.2\% & 73.6\% & 85.6\% & 80.7\% \\
            GoT & 93.2\% & 78.6\% & 89.2\% & 91.0\% & \underline{94.2\%} & 73.3\% & 71.3\% & 70.1\% & 65.2\% & 74.2\% & 84.4\% & 80.4\% \\
            AoT & \underline{95.0\%} & \textbf{83.6\%} & \underline{91.5\%} & \underline{92.6\%} & 94.1\% & \underline{75.1\%} & \underline{74.7\%} & \underline{72.4\%} & \underline{74.7\%} & \underline{75.2\%} & \underline{86.9\%} & 83.3\% \\
            \textbf{Agent-GWO} & \textbf{95.9\%} & \underline{80.2\%} & \textbf{92.3\%} & \textbf{95.3\%} & \textbf{95.5\%} & \textbf{75.9\%} & \textbf{75.3\%} & \textbf{73.9\%} & \textbf{75.8\%} & \textbf{76.4\%} & \textbf{87.5\%} & \textbf{84.0\%} \\
            \midrule
            
            \rowcolor{lightbeige} 
            \multicolumn{13}{c}{\textbf{Qwen2.5-Coder-7B-Instruct}} \\  
            \midrule
            CoT & 77.5\% & 65.8\% & 82.7\% & 84.6\% & 87.3\% & 60.9\% & 55.4\% & 47.4\% & 31.3\% & 20.4\% & 70.4\% & 62.2\% \\
            CoT-SC/n=5 & 80.0\% & 67.3\% & 83.9\% & 86.8\% & 89.8\% & 62.0\% & 56.1\% & 49.1\% & 32.7\% & 20.8\% & 73.9\% & 63.9\% \\
            Self-Refine & 79.0\% & 70.4\% & 83.3\% & 86.4\% & 87.6\% & 61.2\% & 55.8\% & 48.6\% & 31.9\% & 20.8\% & 74.6\% & 63.6\% \\
            AFlow & 84.0\% & 71.9\% & 86.0\% & 86.8\% & 90.1\% & 61.8\% & 56.5\% & 52.5\% & 34.0\% & 22.7\% & 77.0\% & 65.8\% \\
            ToT & 83.6\% & 71.3\% & 85.6\% & 83.0\% & 89.8\% & 61.5\% & \underline{56.9\%} & 55.0\% & 34.5\% & 23.0\% & 77.8\% & 65.6\% \\
            GoT & 84.8\% & 71.7\% & 87.5\% & 85.1\% & 92.5\% & 62.9\% & 56.8\% & 53.4\% & 34.7\% & 23.8\% & 76.8\% & 66.4\% \\
            AoT & \underline{88.5\%} & \underline{72.0\%} & \underline{90.0\%} & \underline{88.4\%} & \textbf{94.1\%} & \underline{63.0\%} & 56.8\% & \underline{57.0\%} & \underline{36.2\%} & \underline{26.4\%} & \underline{78.8\%} & 68.3\% \\
            \textbf{Agent-GWO} & \textbf{89.1\%} & \textbf{74.1\%} & \textbf{90.1\%} & \textbf{93.3\%} & \underline{93.3\%} & \textbf{63.3\%} & \textbf{58.3\%} & \textbf{58.8\%} & \textbf{37.1\%} & \textbf{27.8\%} & \textbf{83.5\%} & \textbf{69.9\%} \\
            \midrule
            
            \rowcolor{lightbeige} 
            \multicolumn{13}{c}{\textbf{Gemma-3-12b-it}} \\ 
            \midrule
            CoT & 83.5\% & 72.8\% & 79.3\% & 82.7\% & 91.0\% & 69.2\% & 68.4\% & 64.9\% & 78.0\% & 49.3\% & 75.1\% & 74.0\% \\
            CoT-SC/n=5 & 85.9\% & 74.6\% & 80.8\% & 85.1\% & 93.2\% & 71.5\% & 70.4\% & 66.5\% & 80.0\% & 52.0\% & 77.0\% & 76.1\% \\
            Self-Refine & 84.6\% & 77.6\% & 80.3\% & 87.5\% & 89.2\% & 70.2\% & 69.1\% & 65.2\% & 79.2\% & 50.6\% & 79.6\% & 75.7\% \\
            AFlow & 89.0\% & 79.0\% & 85.2\% & \underline{89.0\%} & 92.2\% & 73.8\% & 70.8\% & 66.0\% & 81.3\% & 53.0\% & 82.6\% & 78.4\% \\
            ToT & 88.4\% & 79.5\% & 83.9\% & 88.2\% & 91.8\% & 72.6\% & 70.6\% & 68.4\% & 81.0\% & 52.6\% & 83.5\% & 78.2\% \\
            GoT & 89.2\% & \underline{80.3\%} & 86.8\% & 86.8\% & \underline{93.9\%} & 74.5\% & 71.6\% & 68.0\% & 81.5\% & 53.7\% & 82.4\% & 79.0\% \\
            AoT & \underline{91.4\%} & 79.6\% & \textbf{91.8\%} & 87.5\% & 93.5\% & \underline{76.7\%} & \underline{71.9\%} & \underline{68.7\%} & \underline{82.0\%} & \underline{54.8\%} & \underline{84.0\%} & 80.2\% \\
            \textbf{Agent-GWO} & \textbf{92.8\%} & \textbf{82.1\%} & \underline{90.9\%} & \textbf{95.9\%} & \textbf{94.2\%} & \textbf{78.5\%} & \textbf{72.7\%} & \textbf{70.4\%} & \textbf{84.5\%} & \textbf{56.9\%} & \textbf{88.0\%} & \textbf{82.4\%} \\
            \bottomrule
        \end{tabular}
    }
\caption{Performance comparison of Agent-GWO against baselines across math and hybrid reasoning tasks. Bold indicates the best result, and \underline{underlined} indicates the second-best result.}
\label{tab:table1}
\end{table*}

\section{Experiments}
\label{Experiments}

In this section, we conduct a systematic evaluation of the proposed Agent-GWO framework to verify its effectiveness on complex reasoning tasks. Our evaluation covers the reasoning performance (Sec~\ref{sec:app_eval}), adaptability performance to CoT (Sec~\ref{sec:adapt_eval}), and ablation study (Sec~\ref{sec:ablation}).

\subsection{Experimental Setup}
\subsubsection{Datasets}
We evaluate the proposed Agent-GWO framework on a comprehensive suite of 11 benchmarks spanning two complementary categories: mathematical reasoning and hybrid reasoning. The mathematical reasoning set includes GSM8K~\cite{cobbe2021training}, MATH~\cite{hendrycks2021measuring}, SVAMP~\cite{patel2021nlp}, MultiArith~\cite{roy2015solving}, and ASDiv~\cite{miao2020diverse}. The hybrid reasoning set includes AQUA~\cite{ling2017program}, MMLU~\cite{hendrycks2021measuring}, BBH~\cite{suzgun2023challenging}, Date~\cite{suzgun2023challenging}, CLUTRR~\cite{sinha2019clutrr}, and MATH\_MIX.
Collectively, these benchmarks span numerical computation (arithmetic, algebra, geometry, and combinatorics) and broader reasoning skills, including logical inference, knowledge/commonsense recall, temporal reasoning, and multi-step relational chaining. This diversity provides a robust testbed for evaluating GWO on multi-hop, cross-domain, and constraint-heavy reasoning tasks.

\subsubsection{Baselines}
 We compare Agent-GWO with seven representative reasoning-enhancement baselines that can be evaluated under the same backbone setting for a fair comparison: Chain-of-Thought prompting (CoT)~\cite{wei2022chain}, its self-consistency variant CoT-SC with $n{=}5$ samples~\cite{wang2022self}, iterative self-feedback refinement (Self-Refine)~\cite{madaan2023self}, the feedback-driven workflow method AFlow~\cite{zhang2024aflow}, and structured reasoning paradigms including Tree-of-Thought (ToT)~\cite{yao2023tree}, Graph-of-Thought (GoT)~\cite{besta2024graph}, and Atom-of-Thought (AoT)~\cite{teng2025atom}. To assess robustness and generalization across model families and scales, we conduct experiments with five LLM backbones: GPT-4o-mini, GPT-4.1-mini, GPT-4.1-nano, Qwen2.5-Coder-7B-Instruct, and Gemma-3-12b-it.

\subsubsection{Implementation Details}

Following established practices in workflow optimization and inference-time search~\cite{zhang2024aflow,saad2024archon,hu2024automated}, we adopt a strict optimization--evaluation separation protocol. For benchmarks with official splits, including GSM8K, SVAMP, MultiArith, ASDiv, AQUA, MMLU, and CLUTRR, optimization is performed exclusively on the training/dev side to compute exact-match fitness and update prompt--configuration pairs, while the test split is used only for the final evaluation. For MATH, we follow~\cite{hong2025data} and evaluate on 617 Level-5 problems across four representative categories; optimization follows the same protocol using labeled data from the official training side, ensuring that no test labels are used for feedback. For benchmarks without official train/dev splits, including BBH, Date, and MATH\_MIX, we follow prior workflow-optimization practice~\cite{zhang2024aflow,saad2024archon,hu2024automated} and perform a fixed-seed random hold-out split, where the validation and test portions are constructed in a 1:4 ratio. The validation portion serves as the optimization pool for fitness computation, while the test portion is strictly held out for final evaluation.

Unless otherwise specified, we use the default GWO configuration with $n=5$ agents and $K=10$ optimization iterations. This setting provides a practical trade-off between computational budget and performance gains, and is applied consistently across datasets and backbones to ensure fair comparison. Empirical support for this configuration is provided in the ablation study (Section~\ref{sec:ablation}).

\begin{table*}[ht!]
    \centering
    \renewcommand{\arraystretch}{1} 
    
    \resizebox{\textwidth}{!}{
        \begin{tabular}{l|c|c|c|c|c|c|c|c|c|c|c|c} 
            \toprule
            
            \multirow{2}{*}{\textbf{Model / Method}} & 
            \multicolumn{6}{c}{\textbf{Math Reasoning Tasks}} & 
            \multicolumn{5}{c}{\textbf{Hybrid Reasoning Tasks}} &
            \multirow{2}{*}{\textbf{Avg. Increase}} \\ 
            \cmidrule(lr){2-7} \cmidrule(lr){8-12}
            
             & \textbf{GSM8K} & \textbf{MATH} & \textbf{SVAMP} & \textbf{MultiArith} & \textbf{ASDiv} & \textbf{AQUA} 
             & \textbf{MMLU} & \textbf{BBH} & \textbf{Date} & \textbf{CLUTRR} & \textbf{MATH\_MIX} & \\ 
            \midrule

            \rowcolor{lightbeige} 
            \multicolumn{13}{c}{\textbf{GPT-4o-mini}} \\ 
            \midrule
            CoT & 85.4\% & 74.8\% & 84.7\% & 89.5\% & 92.3\% & 65.3\% & 63.4\% & 66.6\% & 52.1\% & 66.2\% & 80.1\% & - \\
            CoT+GWO/n=5     & \textbf{95.1\%} & \textbf{79.9\%} & \textbf{92.4\%} & \textbf{99.5\%} & \textbf{93.8\%} & \textbf{76.1\%} & \textbf{73.7\%} & \textbf{71.3\%} & \textbf{77.1\%} & \textbf{74.9\%} & \textbf{85.0\%} & \textbf{$\uparrow$ 8.9\%} \\
            CoT+GWO/n=6     & \textbf{96.5\%} & \textbf{81.5\%} & \textbf{92.9\%} & \textbf{99.8\%} & \textbf{94.5\%} & \textbf{76.8\%} & \textbf{74.9\%} & \textbf{72.2\%} & \textbf{78.3\%} & \textbf{75.8\%} & \textbf{95.1\%} & \textbf{$\uparrow$ 10.7\%} \\
            \midrule
            
            \rowcolor{lightbeige} 
            \multicolumn{13}{c}{\textbf{GPT-4.1-mini}} \\  
            \midrule
            CoT             & 88.2\% & 79.8\% & 86.1\% & 99.0\% & 91.7\% & 67.5\% & 66.9\% & 69.5\% & 54.9\% & 71.2\% & 82.5\% & - \\
            CoT+GWO/n=5     & \textbf{97.2\%} & \textbf{83.6\%} & \textbf{93.6\%} & \textbf{99.8\%} & \textbf{94.2\%} & \textbf{79.3\%} & \textbf{78.9\%} & \textbf{77.2\%} & \textbf{79.3\%} & \textbf{78.8\%} & \textbf{85.0\%} & \textbf{$\uparrow$ 8.1\%} \\
            CoT+GWO/n=6     & \textbf{98.3\%} & \textbf{84.3\%} & \textbf{94.8\%} & \textbf{99.9\%} & \textbf{94.8\%} & \textbf{80.4\%} & \textbf{79.8\%} & \textbf{78.3\%} & \textbf{80.5\%} & \textbf{79.5\%} & \textbf{97.4\%} & \textbf{$\uparrow$ 10.1\%} \\
            \midrule
            
            \rowcolor{lightbeige} 
            \multicolumn{13}{c}{\textbf{GPT-4.1-nano}} \\  
            \midrule
            CoT             & 83.8\% & 74.3\% & 81.1\% & 98.8\% & 89.6\% & 64.7\% & 61.3\% & 65.7\% & 52.1\% & 64.2\% & 80.0\% & - \\
            CoT+GWO/n=5     & \textbf{93.1\%} & \textbf{80.1\%} & \textbf{92.3\%} & \textbf{99.4\%} & \textbf{93.1\%} & \textbf{77.1\%} & \textbf{74.1\%} & \textbf{70.3\%} & \textbf{77.8\%} & \textbf{72.3\%} & \textbf{83.0\%} & \textbf{$\uparrow$ 8.8\%} \\
            CoT+GWO/n=6     & \textbf{94.2\%} & \textbf{81.2\%} & \textbf{93.8\%} & \textbf{99.7\%} & \textbf{94.0\%} & \textbf{78.2\%} & \textbf{75.1\%} & \textbf{71.2\%} & \textbf{79.3\%} & \textbf{73.5\%} & \textbf{93.5\%} & \textbf{$\uparrow$ 10.7\%} \\
            \midrule
            
            \rowcolor{lightbeige} 
            \multicolumn{13}{c}{\textbf{Qwen2.5-Coder-7B-Instruct}} \\  
            \midrule
            CoT & 77.5\% & 65.8\% & 82.7\% & 84.6\% & 87.3\% & 60.9\% & 55.4\% & 47.4\% & 31.3\% & 20.4\% & 70.4\% & - \\
            CoT+GWO/n=5     & \textbf{89.7\%} & \textbf{72.8\%} & \textbf{90.6\%} & \textbf{97.7\%} & \textbf{90.5\%} & \textbf{62.5\%} & \textbf{58.5\%} & \textbf{54.8\%} & \textbf{37.7\%} & \textbf{28.1\%} & \textbf{75.5\%} & \textbf{$\uparrow$ 6.8\%} \\
            CoT+GWO/n=6     & \textbf{90.6\%} & \textbf{73.8\%} & \textbf{91.5\%} & \textbf{98.1\%} & \textbf{92.0\%} & \textbf{63.1\%} & \textbf{59.1\%} & \textbf{55.6\%} & \textbf{39.2\%} & \textbf{26.1\%} & \textbf{90.2\%} & \textbf{$\uparrow$ 8.7\%} \\
            \midrule
            
            \rowcolor{lightbeige} 
            \multicolumn{13}{c}{\textbf{Gemma-3-12b-it}} \\ 
            \midrule
            CoT & 83.5\% & 72.8\% & 79.3\% & 82.7\% & 91.0\% & 69.2\% & 68.4\% & 64.9\% & 78.0\% & 49.3\% & 75.1\% & - \\
            CoT+GWO/n=5     & \textbf{93.4\%} & \textbf{80.7\%} & \textbf{91.3\%} & \textbf{96.2\%} & \textbf{92.1\%} & \textbf{79.3\%} & \textbf{73.1\%} & \textbf{68.1\%} & \textbf{85.3\%} & \textbf{53.5\%} & \textbf{81.5\%} & \textbf{$\uparrow$ 7.3\%} \\
            CoT+GWO/n=6     & \textbf{94.3\%} & \textbf{81.3\%} & \textbf{92.4\%} & \textbf{96.8\%} & \textbf{93.1\%} & \textbf{80.5\%} & \textbf{74.2\%} & \textbf{68.8\%} & \textbf{86.7\%} & \textbf{54.4\%} & \textbf{94.3\%} & \textbf{$\uparrow$ 9.3\%} \\
            
            \bottomrule
        \end{tabular}
    }
\caption{Combined evaluation of adaptability of GWO to CoT on math and hybrid reasoning tasks.}
\label{tab:combined_adaptability}
\end{table*}

\subsection{Performance Evaluation}
\label{sec:app_eval}

Table~\ref{tab:table1} reports a systematic comparison of Agent-GWO against seven representative reasoning-enhancement baselines (CoT, CoT-SC, Self-Refine, AFlow, ToT, GoT, and AoT) across eleven benchmarks under three LLM backbones. Overall, Agent-GWO achieves the best average performance (Avg) on all backbones, and attains the best or second-best results on most individual tasks, demonstrating consistent stability and strong competitiveness across settings. In contrast, methods such as CoT and Self-Refine yield more limited overall gains on complex mathematical and hybrid reasoning benchmarks, and their effectiveness is more sensitive to the specific task and backbone configuration, leading to larger performance fluctuations. These results also highlight the benefit of Agent-GWO's collaborative optimization: by leveraging dynamic multi-agent interactions to iteratively calibrate and consolidate the reasoning process, it mitigates error accumulation in long-chain inference and improves robustness and generalization in challenging reasoning scenarios. More details of the evaluation pipeline are provided in Appendix~\ref{Performanc1}.

\subsection{Adaptability Evaluation}
\label{sec:adapt_eval}

To further assess the adaptability and generalization of Agent-GWO, we integrate it with the mainstream reasoning paradigm CoT and evaluate the combined method across a broad set of mathematical and hybrid reasoning benchmarks. Specifically, we treat standard CoT prompting as the baseline and measure the accuracy gains obtained by incorporating Agent-GWO into the CoT pipeline. All runs are conducted independently, and accuracy is computed on the official validation sets of each benchmark. The aggregated results are reported in Table~\ref{tab:combined_adaptability}.

Across six math and five hybrid reasoning benchmarks, CoT+GWO yields consistent and monotonic gains over vanilla CoT for all evaluated backbones, with larger improvements as the iteration budget $n$ increases. For GPT-4o-mini, CoT+GWO boosts GSM8K/MATH/AQUA from 85.4/74.8/65.3 to 95.1/79.9/76.1 with $n{=}5$, and further to 96.5/81.5/76.8 with $n{=}6$. When averaging over all 11 benchmarks, this corresponds to an absolute accuracy improvement of 8.9 percentage points for $n{=}5$ and 10.7 percentage points for $n{=}6$ relative to CoT. This consistent trend extends across various model families and scales, including Gemma and Qwen, as well as hybrid reasoning tasks. The gains are particularly pronounced on the MATH\_MIX benchmark, where Qwen2.5-Coder-7B-Instruct achieved a substantial accuracy increase from 70.4\% to 90.2\%.

Overall, Table~\ref{tab:combined_adaptability} shows that integrating Agent-GWO with CoT consistently improves reasoning performance across diverse benchmarks and model architectures, spanning mathematical, logical, temporal, and mixed-domain tasks. The consistent gains observed on both lightweight and stronger backbones highlight the generality and transferability of the proposed inference-time optimization framework, positioning Agent-GWO as an effective complement to standard CoT prompting for enhancing reasoning robustness and task alignment.

\begin{figure*}[ht!]
    \centering
    \begin{minipage}[b]{0.48\linewidth}
        \centering
        \includegraphics[width=\linewidth]{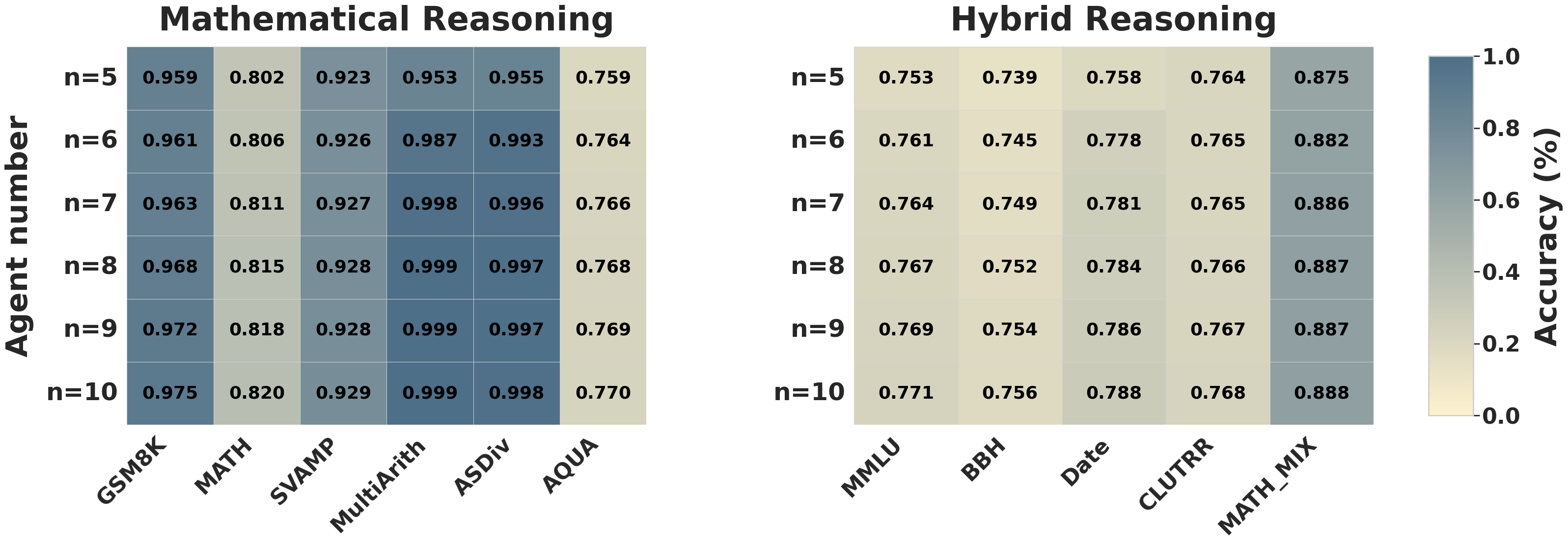}
        \caption{Accuracy over varying number of agents on mathematical (right) and hybrid (left) reasoning tasks.}
        \label{fig:temp-plot19}
    \end{minipage}
    \hfill 
    \begin{minipage}[b]{0.48\linewidth}
        \centering
        \includegraphics[width=\linewidth]{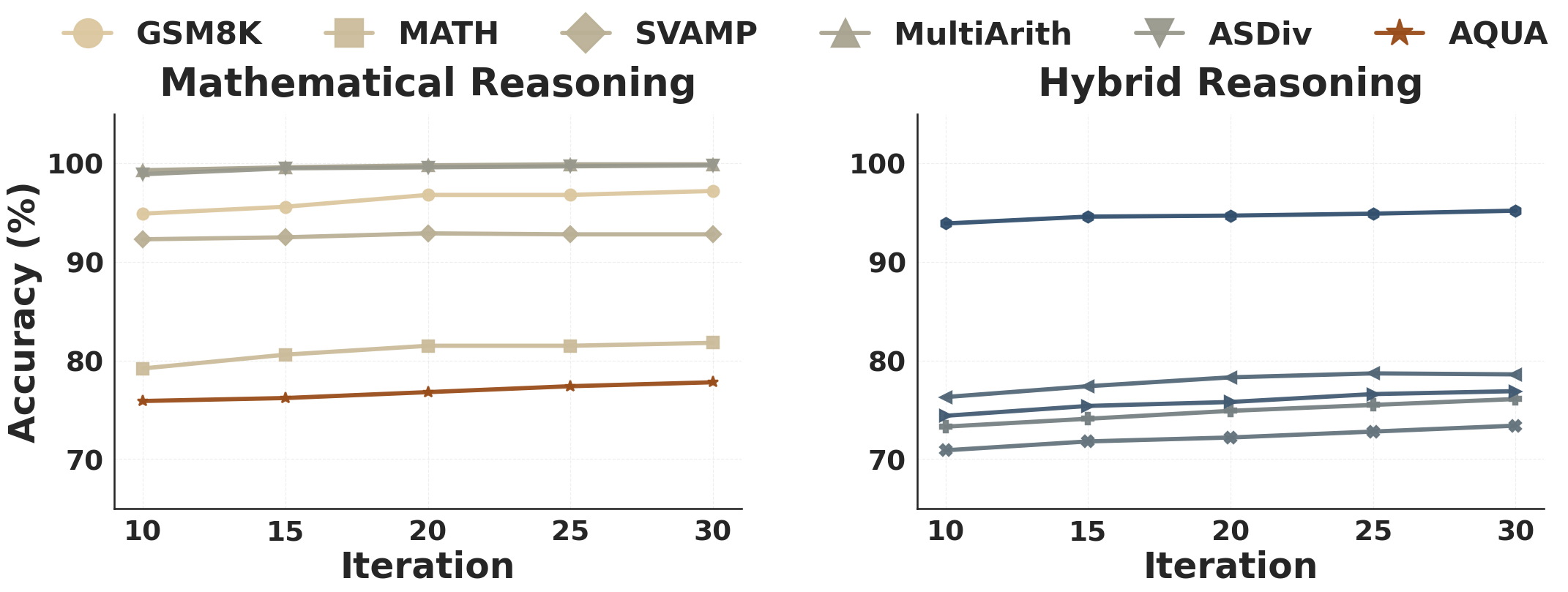}
        \caption{Accuracy over iteration for math (left) and hybrid (right) reasoning tasks (fixed five agents).}
        \label{fig:concept1}
    \end{minipage}
\end{figure*}

\subsection{Hyper-parameter Sensitivity Analysis}
\label{sec:sensitivity}

To systematically evaluate the robustness and stability of Agent-GWO under decoding-level configurations, we conduct a sensitivity analysis on five key agent-level hyperparameters: temperature $t_j$, top-$p$ threshold $p_j$, frequency penalty $f_j$, presence penalty $e_j$, and maximum generation length $m_j$. These hyperparameters regulate the stochasticity, diversity, repetition, novelty, and reasoning budget of LLM-based agents, and collectively shape agent behaviors during the optimization process. For continuous decoding parameters ($t_j, p_j, f_j, e_j$), each agent samples its configuration from a normal distribution $\mathcal{N}(\mu,\sigma^2)$, followed by clipping to a valid range. This strategy introduces controlled variability across agents while preventing unstable or degenerate decoding settings. The maximum generation length $m_j$ is treated separately and sampled uniformly from predefined intervals, enabling explicit control over the available reasoning budget and inference cost.

\begin{table}[t!]
\centering
\renewcommand{\arraystretch}{1}
\scalebox{0.8}{
\begin{tabular}{c c c c}
\toprule
\rowcolor{lightbeige}
\textbf{Parameter} & \textbf{Sampling Rule} & \textbf{Perf.(\%)} & \textbf{Cost (\$)} \\
\midrule

\multirow{4}{*}{$t_j$}
& (0.2, 0.1) & $\downarrow$1.0--1.5\% & $\downarrow$3--5\% \\
& (0.4, 0.1) & $\downarrow$0.3--0.6\% & $\downarrow$1--2\% \\
& (0.6, 0.1) & baseline               & baseline \\
& (0.8, 0.1) & $\downarrow$0.5--1.0\% & $\uparrow$4--7\% \\
\midrule

\multirow{4}{*}{$p_j$}
& (0.2, 0.1) & $\downarrow$0.8--1.3\% & $\downarrow$2--4\% \\
& (0.4, 0.1) & $\downarrow$0.2--0.5\% & $\downarrow$1--2\% \\
& (0.6, 0.1) & baseline               & baseline \\
& (0.8, 0.1) & $\uparrow$0--0.3\%          & $\uparrow$2--4\% \\
\midrule

\multirow{4}{*}{$f_j$}
& (0.0, 0.1) & $\downarrow$0.6--1.0\% & $\uparrow$4--6\% \\
& (0.4, 0.1) & $\downarrow$0.2--0.4\% & $\uparrow$1--2\% \\
& (0.6, 0.1) & baseline               & baseline \\
& (0.8, 0.1) & $\pm$0--1\% & $\downarrow$1--3\% \\
\midrule

\multirow{4}{*}{$e_j$}
& (0.0, 0.1) & $\downarrow$0.4--0.8\% & $\uparrow$2--4\% \\
& (0.4, 0.1) & $\downarrow$0.1--0.3\% & $\uparrow$1--2\% \\
& (0.6, 0.1) & baseline               & baseline \\
& (0.8, 0.1) & $\downarrow$0.5--1.0\% & $\uparrow$3--6\% \\
\midrule

\multirow{4}{*}{$m_j$}
& [1024,1274]  & $\downarrow$0.5--1.5\% & $\downarrow$5--12\% \\
& [1274,1524]  & baseline               & baseline \\
& [1524,1774]  & $\uparrow$0--0.4\%     & $\uparrow$3--8\% \\
& [1774,2024]  & $\uparrow$0--0.8\%     & $\uparrow$6--15\% \\
\bottomrule
\end{tabular}}
\caption{Sensitivity analysis of agent-level decoding hyperparameters. Entries report relative performance and cost changes compared to the baseline configuration used in our framework.}
\label{tab:sensitivity_mu_sigma}
\end{table}

Table~\ref{tab:sensitivity_mu_sigma} reports the relative changes in task performance and inference cost under different decoding hyperparameter settings, normalized to the default configuration. All sensitivity experiments are conducted on the gpt-4o-mini model using the GSM8K dataset. Empirical results show that Agent-GWO is largely insensitive to moderate variations in decoding hyperparameters. Across a wide range of $(\mu,\sigma)$ settings and token-length intervals, performance fluctuations remain limited, typically within $\pm 1.5\%$. Moderate values of $t_j$ and $p_j$ consistently yield stable performance, while extreme settings tend to degrade accuracy due to insufficient diversity or excessive randomness. The penalty terms $f_j$ and $e_j$ mainly affect output redundancy and repetition, with limited influence on accuracy but observable impact on inference cost. Among all parameters, $m_j$ exhibits the most direct trade-off between performance and computational cost.

Overall, this analysis suggests that Agent-GWO does not rely on delicate hyperparameter tuning to achieve stable performance, supporting its practical reliability and reproducibility across different decoding configurations.

\subsection{Ablation Study}
\label{sec:ablation}

\paragraph{Impact of Agent Population and Iteration.}
To analyze the contribution of the number of agents and iterations in the GWO framework, we conduct an ablation study on GPT-4o-mini across multiple datasets. We first examine the effect of varying the number of agents, with $n \in \{5, 6, 7, 8, 9, 10\}$. As shown in Figure~\ref{fig:temp-plot19}, accuracy on both mathematical (GSM8K, MATH, SVAMP, MultiArith, ASDiv, AQUA) and hybrid reasoning benchmarks (MMLU, BBH, Date, CLUTRR, MATH\_MIX) generally improves as the number of agents increases. For example, GSM8K accuracy rises from 95.9\% at $n=5$ to 97.5\% at $n=10$, while MMLU improves from 75.3\% to 77.1\%. The gains, however, are non-linear and diminish as $n$ grows, indicating that moderate increases in agent count yield limited additional benefits and should be weighed against computational cost.

We further study the effect of the number of optimization iterations while fixing the number of agents at $n=5$. As illustrated in Figure~\ref{fig:concept1}, increasing the number of iterations from 10 to 30 consistently improves performance across both mathematical and hybrid reasoning tasks. This trend suggests that additional iterations enable more thorough exploration of the solution space and progressive refinement of reasoning strategies, leading to more stable convergence. At the dataset level, most benchmarks exhibit monotonic or near-monotonic improvements as the iteration count increases, underscoring the importance of iterative optimization in GWO.

\begin{table}[!t]
\scalebox{0.8}{
\begin{tabular}{llccc}
\toprule
\rowcolor{lightbeige}
\textbf{Elite Size} & \textbf{Elite} & \textbf{GSM8K} & \textbf{MATH} & \textbf{MMLU} \\
\midrule
$m=1$ & $\alpha$ & 92.4 & 76.8 & 72.0 \\
$m=2$ & $\alpha, \beta$ & 94.8 & 79.1 & 74.2 \\
$m=3$ & $\alpha, \beta, \delta$ & \textbf{95.9} & \textbf{80.2} & \textbf{75.3} \\
$m=4$ & $\alpha, \beta, \delta, \omega_1$ & 95.0 & 79.4 & 74.6 \\
$m=5$ & $\alpha, \beta, \delta, \omega_1, \omega_2$ & 94.1 & 78.5 & 73.8 \\
\bottomrule
\end{tabular}}
\caption{Ablation on Elite Set Size. Results under different elite set sizes $m$. $\alpha$, $\beta$, and $\delta$ denote the top three leaders in standard GWO, while $\omega_1$ and $\omega_2$ denote the 4th- and 5th-ranked agents by fitness, respectively.}
\label{tab:m_ablation}
\end{table}

\paragraph{Impact of the Number of Elite Agents.}
To understand how the diversity of the guidance signal affects the search process, we investigate the sensitivity of our method to the number of elite agents, denoted as $m$. In standard GWO, the search is guided by the top-three leaders ($\alpha$, $\beta$, and $\delta$). To verify if this configuration is optimal for our discrete search space, we vary $m$ from $1$ to $5$. As illustrated in Table~\ref{tab:m_ablation}, we observe a clear inverted U-shaped trend across the GSM8K, MATH, and MMLU benchmarks. Increasing $m$ from $1$ to $3$ steadily improves performance, as the added leader diversity helps the algorithm avoid premature convergence. However, when $m > 3$, the inclusion of lower-ranked agents (e.g., $\omega_1$ and $\omega_2$) mixes sub-optimal prompts into the update step. This dilutes the strong guidance signal from the top leaders and reduces the overall selection pressure, leading to a slight drop in accuracy. Consequently, we maintain $m=3$ as the default configuration.

Overall, the number of agents, iterations, and elite set size all affect the effectiveness of GWO. Increasing the agent population enhances global exploration, while additional iterations improve convergence stability and reduce variance, and a proper elite set size helps balance guidance diversity and selection pressure. Together, these factors support the scalability of the GWO framework, enabling it to better leverage the reasoning potential of large models across various tasks.

\section{Conclusion}
In this paper, we propose Agent-GWO, a dynamic prompt optimization framework tailored for complex reasoning tasks. By unifying prompt templates and decoding configurations into iteratively optimizable agent states, we design a hierarchical collaborative mechanism that enables agents to progressively refine strategies through mutual learning over multiple iterations. This swarm-based perspective facilitates stable convergence while effectively maintaining exploration capabilities, allowing for the automatic discovery of robust, task-aligned reasoning configurations that are directly deployable for inference. Extensive experiments across diverse reasoning benchmarks and varying computational budgets demonstrate that Agent-GWO consistently outperforms both static prompting baselines and existing automatic prompt optimization methods in terms of accuracy and stability.

\section*{Limitations}
Agent-GWO relies on parallel exploration with multiple agents and iterative updates at test time; accordingly, it incurs a higher inference-time compute budget than single-pass prompting and should be configured with appropriate $(n,K)$ under practical compute and latency constraints. In addition, our empirical study primarily focuses on mathematical and hybrid reasoning benchmarks. While the proposed framework is broadly applicable in form, further evaluations on a wider range of domain-specific and real-world tasks are needed to more fully characterize its effectiveness and operating boundaries.

In future work, we will further improve the efficiency and generalization of the framework. One important direction is to develop more compute-aware agent scheduling and adaptive update strategies to reduce the cost of multi-agent exploration while maintaining optimization quality. Another direction is to extend Agent-GWO to broader application scenarios, including domain-specific reasoning tasks and open-ended real-world tasks, so as to more comprehensively evaluate its robustness and transferability.

\section*{Acknowledgments}
This work was supported by the National Natural Science Foundation of China (NSFC) under the General Program (Grant No. 62572104).

\bibliography{custom}

\appendix

\newpage
\section{Appendix}
\label{sec:appendix}

\subsection{Preliminary Knowledge for Grey Wolf Optimizer}
\label{sec:gwo_details}
\begin{figure*}[h]
    \centering
    \includegraphics[width=1\linewidth]{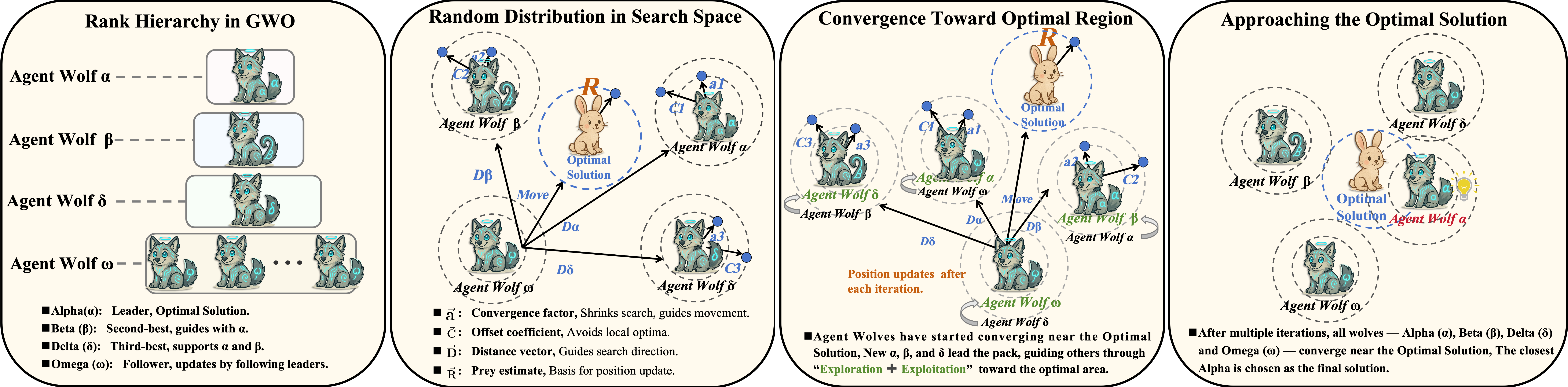}
    \caption{Illustration of GWO algorithm.}
    \label{fig:concept11}
\end{figure*}

\textbf{Grey Wolf Optimizer}. GWO is a population-based metaheuristic inspired by the social hierarchy and hunting behavior of grey wolves. Designed for solving continuous optimization problems, GWO is characterized by its simplicity and minimal reliance on hyperparameters. In this algorithm, a population of $N$ wolves is maintained, where each wolf represents a candidate solution $\mathbf{X}_i = (x_i^1, x_i^2, \dots, x_i^D)$ in a $D$-dimensional search space. The objective is to optimize a target function $f(\mathbf{X})$. GWO mimics the hierarchical structure of grey wolves, dividing the population into four roles: $\alpha$ (leader), $\beta$ (second-in-command), $\delta$ (subordinate), and $\omega$ (follower). During the optimization process, the $\alpha$, $\beta$, and $\delta$ wolves correspond to the top three solutions and are responsible for guiding the remaining $\omega$ wolves toward promising regions in the search space, as illustrated in Figure~\ref{fig:concept11}.

The core mechanism of the GWO is inspired by the encircling behavior exhibited during hunting. 
In each iteration, wolves estimate their distance from the prey (i.e., the current best solution) as 
$ \mathbf{D}(t) = \left| \mathbf{C}(t) \odot \mathbf{X}_p(t) - \mathbf{X}_i(t) \right| $, 
where $\mathbf{X}_p(t)$ denotes the position of the prey at iteration $t$, 
$\mathbf{C}(t) = 2\mathbf{r}_1(t)$, and $\mathbf{r}_1(t)\in [0,1]^D$ is a uniformly distributed random vector.  

Based on $\mathbf{D}(t)$, the position of each wolf is updated by 
$ \mathbf{X}_i(t+1) = \mathbf{X}_p(t) - \mathbf{A}(t) \odot \mathbf{D}(t) $, 
where $\mathbf{A}(t) = 2\,\mathbf{a}(t)\odot \mathbf{r}_2(t) - \mathbf{a}(t)$ and 
$\mathbf{r}_2(t)\in [0,1]^D$ is another random vector. The parameter decreases linearly with iterations as 
$ \mathbf{a}(t) = 2 - 2 \cdot \tfrac{t}{T_{\max}} $, 
where $T_{\max}$ is the maximum number of iterations. This design allows the algorithm to balance exploration in early stages and exploitation in later stages.  

Each $\omega$ wolf also updates its position by referring to the three leading wolves $\alpha$, $\beta$, and $\delta$. 
The distances are computed as 
$ \mathbf{D}_j(t) = \left| \mathbf{C}_j(t) \odot \mathbf{X}_j(t) - \mathbf{X}_i(t) \right| $, $j\in\{\alpha,\beta,\delta\}$, 
and the corresponding candidate updates are 
$ \mathbf{X}_i^{(j)}(t+1) = \mathbf{X}_j(t) - \mathbf{A}_j(t) \odot \mathbf{D}_j(t) $.  
Finally, the new position of wolf $i$ is obtained by averaging the three guided positions: 

\begin{equation}
    \mathbf{X}_i(t+1) = \tfrac{\mathbf{X}_i^{(\alpha)}(t+1)+\mathbf{X}_i^{(\beta)}(t+1)+\mathbf{X}_i^{(\delta)}(t+1)}{3}
\end{equation}

By integrating leader-based guidance with stochastic exploration, the GWO effectively achieves a balance between global search and local exploitation. As iterations progress, the decreasing parameter $ \mathbf{a}(t) $ ensures a smooth transition from exploration to exploitation, enhancing convergence toward the global optimum.

\subsection{LLM-as-a-Judge Protocol}
\label{sec:LLM-judge}

We adopt a controlled \emph{LLM-as-a-Judge} protocol as an auxiliary evaluation mechanism to assess the quality of intermediate reasoning states and state transitions, particularly in settings where explicit gold answers are unavailable.

\paragraph{Evaluation Setup.}
A fixed large language model is used as the judge across all experiments. The judge model, prompt, and decoding configuration are kept identical throughout training and evaluation. The prompt is frozen prior to experimentation and is not adapted based on agent outputs or historical performance, preventing self-referential evaluation.

For each transition, the judge is provided with the current state $Q_i$, the candidate next state $Q_{i+1}$, and their corresponding reasoning trajectories. The judge evaluates whether the transition preserves the original reasoning objective while offering improved reasoning utility.

\paragraph{Evaluation Criteria.}
The judge produces three normalized scores in $[0,1]$:
\[
\bigl(s^{\mathrm{logic}},\; s^{\mathrm{creativity}},\; s^{\mathrm{complete}}\bigr),
\]
corresponding to logical consistency, structural novelty introduced by the transition, and coverage of required reasoning elements, respectively. The judge is explicitly instructed not to rely on superficial cues such as verbosity or stylistic features.

\paragraph{Usage in Different Settings.}
For verifiable benchmarks with gold answers, exact-match accuracy remains the primary fitness signal. Judge-based scores are used only as tie-breakers when multiple agents achieve identical accuracy.  
For non-verifiable tasks, agent fitness is computed as a weighted sum of judge scores:
\[
\mathrm{FITNESS}_j = w_1 s^{\mathrm{logic}} + w_2 s^{\mathrm{creativity}} + w_3 s^{\mathrm{complete}},
\]
with weights $(w_1, w_2, w_3) = (0.5, 0.2, 0.3)$ calibrated on 200 human-labeled samples.

\paragraph{Stability.}
All judge-based scores are averaged over three independent runs with different random seeds to reduce variance and improve ranking stability.

\subsection{Pseudocode}
\label{Pseudocode}

\begin{algorithm}[h]
\caption{Grey Wolf Optimizer (GWO)}
\label{alg:gwo}
\small
\begin{algorithmic}[1]
\Require Population size $N$, dimension $D$, objective $\mathcal{F}$, max iterations $T_{\max}$
\Ensure Best solution $\mathbf{X}^\star$

\Statex \textbf{Initialization}
\State Initialize population $\{\mathbf{X}_i\}_{i=1}^N \subset \mathbb{R}^D$ within bounds
\State $a \gets 2$
\State $\mathbf{X}^\star \gets \arg\min_{\mathbf{X}_i}\, \mathcal{F}(\mathbf{X}_i)$

\Statex \textbf{Iterative Optimization}
\For{$t = 1$ \textbf{to} $T_{\max}$}
    \State Evaluate $\mathcal{F}(\mathbf{X}_i)$ for all $i$
    \State Identify leader wolves $\alpha, \beta, \delta$ (top-3 by fitness)
    \State $a \gets 2 - 2t / T_{\max}$
    \ForAll{$\mathbf{X}_i$}
        \ForAll{$\ell \in \{\alpha, \beta, \delta\}$}
            \State Sample $\mathbf{r}_1, \mathbf{r}_2 \sim \mathcal{U}(0,1)^D$
            \State $\mathbf{A}_\ell \gets 2a \mathbf{r}_1 - a$
            \State $\mathbf{C}_\ell \gets 2 \mathbf{r}_2$
            \State $\mathbf{D}_\ell \gets \lvert \mathbf{C}_\ell \odot \mathbf{X}_\ell - \mathbf{X}_i \rvert$
            \State $\mathbf{X}'_\ell \gets \mathbf{X}_\ell - \mathbf{A}_\ell \odot \mathbf{D}_\ell$
        \EndFor
        \State $\mathbf{X}_i \gets \frac{1}{3}(\mathbf{X}'_\alpha + \mathbf{X}'_\beta + \mathbf{X}'_\delta)$
        \State Project $\mathbf{X}_i$ to feasible bounds
    \EndFor
    \State Update $\mathbf{X}^\star$ if improved
\EndFor

\State \Return $\mathbf{X}^\star$
\end{algorithmic}
\end{algorithm}

\begin{algorithm}[h]
\caption{Agent-Level Iterative Optimization via GWO}
\label{alg:gwo-agent}
\small
\begin{algorithmic}[1]
\Require Dataset $\mathcal{D}$, number of agents $n$, iterations $K$
\Ensure Optimal agent configuration $(\boldsymbol{\eta}^*, \textit{prompt}^*)$

\Statex \textbf{Initialization}
\State Initialize empty agent set $\mathcal{A} \gets \emptyset$
\For{$j = 1$ \textbf{to} $n$}
    \State Sample hyperparameters $\boldsymbol{\eta}_j \sim \mathcal{N}(\boldsymbol{\mu}, \boldsymbol{\sigma}^2)$
    \State $\boldsymbol{\eta}_j \gets \mathrm{Clip}(\boldsymbol{\eta}_j)$
    \State Sample prompt template $\textit{prompt}_j$
    \State $\mathcal{A} \gets \mathcal{A} \cup \{(\boldsymbol{\eta}_j, \textit{prompt}_j)\}$
\EndFor

\Statex \textbf{Iterative Optimization}
\For{$k = 1$ \textbf{to} $K$}
    \ForAll{$(\boldsymbol{\eta}_j, \textit{prompt}_j) \in \mathcal{A}$}
        \State Sample question $q \sim \mathcal{D}$
        \State $(\textit{CoT}_j, \textit{Ans}_j) \gets \mathrm{LLM}(\boldsymbol{\eta}_j, \textit{prompt}_j, q)$
        \State $\textit{Fitness}_j \gets \mathrm{Evaluate}(\textit{Ans}_j, q)$
    \EndFor

    \State Select leader agents $(\alpha, \beta, \delta)$ by fitness
    \State Define sampling weights $\mathbf{w} = (w_\alpha, w_\beta, w_\delta)$, $\sum w = 1$

    \ForAll{$(\boldsymbol{\eta}_j, \textit{prompt}_j) \in \mathcal{A} \setminus \{\alpha,\beta,\delta\}$}
        \State Sample leader $r \sim \{\alpha,\beta,\delta\}$ according to $\mathbf{w}$
        \State $\boldsymbol{\eta}_j \gets \mathrm{Clip}(\mathcal{N}(\boldsymbol{\eta}_r, \sigma^2))$
        \State $\textit{prompt}_j \gets \mathrm{PromptAdaptation}(\textit{prompt}_j, \{\textit{prompt}_\alpha,\textit{prompt}_\beta,\textit{prompt}_\delta\})$
    \EndFor
\EndFor

\Statex \textbf{Output}
\State Extract best agent $(\boldsymbol{\eta}^*, \textit{prompt}^*) \gets (\boldsymbol{\eta}_\alpha, \textit{prompt}_\alpha)$
\State \Return $(\boldsymbol{\eta}^*, \textit{prompt}^*)$
\end{algorithmic}
\end{algorithm}

Algorithm~\ref{alg:gwo} presents the standard GWO: evaluate all candidates, select the top three leaders $(\alpha,\beta,\delta)$, update the rest by aggregating leader-guided proposals with annealed coefficients, project to bounds, and track the best $\mathbf{X}^\star$.

Algorithm~\ref{alg:gwo-agent} instantiates GWO for our setting, where each agent is $(\boldsymbol{\eta}_j,\textit{prompt}_j)$. Each iteration scores agents on sampled questions, selects leaders, and updates non-leaders by following leader hyperparameters (perturb + clip) and adapting prompts via $\mathrm{PromptAdaptation}(\cdot)$, returning the best leader $(\boldsymbol{\eta}^*,\textit{prompt}^*)$.

\subsection{More Experimental Results}
\label{Performanc1}
To further evaluate the effectiveness of our GWO-based prompt optimization, we conduct benchmarks on two standard math reasoning datasets, \textsc{GSM8K} and \textsc{MATH}, using Qwen2.5-Coder-7B-Instruct as the primary backbone. Table~\ref{tab:qwen} reports accuracy (\%) for representative prompt optimization and reasoning baselines. Compared to vanilla CoT and CoT with self-consistency sampling (CoT-SC), GWO yields a substantial improvement, reaching 89.1\% on \textsc{GSM8K} and 72.1\% on \textsc{MATH}. When combined with CoT, the performance is further strengthened, and our best setting (GWO/$n$=6 + CoT) achieves 90.6\% on \textsc{GSM8K} and 73.8\% on \textsc{MATH}, outperforming the listed strong baselines under the same backbone.

In addition, we compare against widely-used GPT-4 prompting/optimization baselines on \textsc{GSM8K} to contextualize the strength of the resulting configuration. As shown in Table~\ref{tab:gpt4}, GWO/$n$=6 + CoT with GPT-4o-mini attains 96.5\% accuracy on \textsc{GSM8K}, which is competitive with multiple optimized prompting methods reported for GPT-4. Overall, these results indicate that GWO consistently improves upon CoT-style prompting across backbones and datasets, and its combination with CoT is complementary, providing a simple yet effective way to obtain stronger and more stable reasoning performance.

\begin{table}[ht!]
\centering
\captionsetup{width=\linewidth, skip=4pt}
\caption{\small Accuracy on GSM8K comparing GPT-4 and GPT-4o-mini.}
\label{tab:gpt4}
\scalebox{0.7 }{ 
\renewcommand{\arraystretch}{1}
\begin{tabular}{@{}l|l|c@{}}
\toprule
\textbf{Method} & \textbf{Model} & \textbf{GSM8K} \\
\midrule
CoMAT~\cite{leang2025comat} & GPT-4 & 93.7 \\
CoT~\cite{wei2022chain} & GPT-4 & 94.5 \\
FCoT~\cite{lyu2023faithful} & GPT-4 & 95.0 \\
MathPrompter~\cite{imani2023mathprompter} & GPT-4 & 95.6 \\
QuaSAR~\cite{radford2018improving} & GPT-4 & 96.5 \\
MathDivide~\cite{srivastava2024mathdivide} & GPT-4 & 96.8 \\
\rowcolor{lightbeige} 
\textbf{GWO/n=6+CoT} & \textbf{GPT-4o-mini} & \textbf{96.5} \\
\bottomrule
\end{tabular}
}
\end{table}

\begin{table}[ht!]
\centering
\captionsetup{width=\linewidth, skip=4pt}
\caption{\small Accuracy (\%) on GSM8K and MATH using Qwen2.5-Coder-7B.}
\label{tab:qwen}
\scalebox{0.75}{
\begin{tabular}{l|c|c|c}
\toprule
\textbf{Method} & \textbf{Base} & \textbf{GSM8K} & \textbf{MATH} \\ 
\midrule
OMI2~\cite{li2025codei} & Qwen2.5 & 84.1 & 72.3 \\
CODEI/O++~\cite{li2025codei} & Qwen2.5 & 85.7 & 72.1 \\
PyEdu~\cite{li2025codei} & Qwen2.5 & 85.8 & 71.4 \\
CODEI/O~\cite{li2025codei} & Qwen2.5 & 86.4 & 71.9 \\
OC-SFT-1~\cite{li2025codei} & Qwen2.5 & 86.7 & 70.9 \\
WI~\cite{li2025codei} & Qwen2.5 & 87.0 & 71.4 \\
WI (Full)~\cite{li2025codei} & Qwen2.5 & 87.0 & 71.1 \\
OMI2 (Full)~\cite{li2025codei} & Qwen2.5 & 88.5 & 73.2 \\
CoT & Qwen2.5 & 77.3 & 69.7 \\
CoT-SC/n=5 & Qwen2.5 & 80.1 & 71.2 \\
GWO & Qwen2.5 & 89.1 & 72.1 \\
GWO+CoT & Qwen2.5 & 89.7 & 72.8 \\
\rowcolor{lightbeige} 
\textbf{GWO/n=6+CoT} & \textbf{Qwen2.5} & \textbf{90.6} & \textbf{73.8} \\
\bottomrule
\end{tabular}
}
\end{table}

\clearpage
\subsection{Notation}
\label{sec:notation}
{\small
\renewcommand{\arraystretch}{1.3}
\setlength{\aboverulesep}{0pt}
\setlength{\belowrulesep}{2pt}
\begin{tabular}{cl}
\toprule
\textbf{Symbol} & \textbf{Definition} \\
\midrule
\multicolumn{2}{l}{\textbf{Grey Wolf Optimizer (GWO) Symbols}} \\
\midrule
$N$            & Population size (number of wolves/solutions) \\
$D$            & Search space dimensionality \\
$\mathbf{X}_i$ & Position of the $i$-th wolf: $\mathbf{X}_i = (x_i^1, x_i^2, \dots, x_i^D)$ \\
$f(\mathbf{X})$ & Objective function to be optimized \\
$T_{\max}$     & Maximum number of iterations \\
$t$            & Current iteration number \\
$\mathbf{X}_p(t)$ & Prey position at iteration $t$ (current best solution) \\
$\mathbf{C}$   & Coefficient vector, $\mathbf{C} = 2\mathbf{r}_1$, $\mathbf{r}_1 \sim \mathcal{U}(0,1)^D$ \\
$\mathbf{D}$   & Distance vector, $\mathbf{D} = |\mathbf{C} \cdot \mathbf{X}_p(t) - \mathbf{X}_i(t)|$ \\
$\mathbf{A}$   & Adaptive coefficient, $\mathbf{A} = 2\mathbf{a} \cdot \mathbf{r}_2 - \mathbf{a}$, $\mathbf{r}_2 \sim \mathcal{U}(0,1)^D$ \\
$\mathbf{a}$   & Linearly decreasing parameter, $\mathbf{a} = 2 - 2 \cdot \frac{t}{T_{\max}}$ \\
$\alpha, \beta, \delta$ & The top three wolves (leaders) in GWO \\
$\omega$       & The remaining wolves (followers) \\
$\mathbf{D}_j$ & Distance from wolf $i$ to leader $j$: $\mathbf{D}_j = |\mathbf{C}_j \cdot \mathbf{X}_j - \mathbf{X}_i|$, $j \in \{\alpha, \beta, \delta\}$ \\
$\mathbf{A}_j, \mathbf{C}_j$ & Adaptive/random coefficients for leader $j$ \\
$\mathbf{X}_j'$ & Updated position guided by leader $j$ \\
$\mathcal{F}$  & Fitness function \\
\midrule
\multicolumn{2}{l}{\textbf{Agent Symbols}} \\
\midrule
$n$            & Number of agents \\
$\text{Agent}_j$ & The $j$-th agent, $\text{Agent}_j = \{\text{LLM}_j, \text{prompt}_j\}$ \\
$\text{LLM}_j$ & Large language model of agent $j$, $\text{LLM}_j = \{\boldsymbol{\eta}_j, \boldsymbol{\theta}\}$ \\
$\boldsymbol{\theta}$ & Shared model parameter set of LLMs \\
$\boldsymbol{\eta}_j$ & Agent-specific hyperparameter set: $\{T_j, p_j, F_j, E_j, M_j\}$ \\
$T_j$          & Temperature hyperparameter for agent $j$ \\
$p_j$          & Top-$p$ threshold for agent $j$ \\
$F_j$          & Frequency penalty for agent $j$ \\
$E_j$          & Presence penalty for agent $j$ \\
$M_j$          & Maximum token length for agent $j$ \\
$\text{clip}(x, [a, b])$ & Clipping function: $\max(a, \min(x, b))$ \\
$\mathcal{N}(\mu, \sigma^2)$ & Normal distribution with mean $\mu$ and variance $\sigma^2$ \\
$\mathcal{U}(a, b)$ & Uniform distribution over $[a, b]$ \\
$\mathcal{M}$  & Discrete set of possible maximum token lengths \\
$c_M$          & Constant value for maximum token length (fixed-length tasks) \\
\midrule
\multicolumn{2}{l}{\textbf{Data and Task Symbols}} \\
\midrule
$\mathcal{D}$  & Reasoning problem dataset, $\mathcal{D} = \{q_1, q_2, \dots, q_N\}$ \\
$q$            & A single question from dataset $\mathcal{D}$ \\
$\text{CoT}_j$ & Chain of Thought generated by agent $j$ \\
$\text{Answer}_j$ & Final answer generated by agent $j$ \\
$f(\text{Agent}_j, q)$ & Output of agent $j$ on $q$: $(\text{CoT}_j, \text{Answer}_j)$ \\
\midrule
\multicolumn{2}{l}{\textbf{Optimization Process Symbols}} \\
\midrule
$K$            & Number of optimization iterations in multi-agent GWO \\
$\mathcal{A}$  & Population set of agents \\
$\mathbf{w}$   & Weight vector for leaders: $\mathbf{w} = \{w_\alpha, w_\beta, w_\delta\}$, $w_\alpha > w_\beta > w_\delta$, $\sum w = 1$ \\
$\text{promptAdaptation}$ & prompt template adaptation function \\
$(\boldsymbol{\eta}^*, \textit{prompt}^*)$ & Optimal hyperparameters and prompt template found \\
\bottomrule
\end{tabular}
}

\clearpage
\subsection{Training Examples}

\subsubsection{Forward Inference Examples}

This example presents a math word problem about recycling cans. The problem describes that Tiffany was collecting cans for recycling. On Monday, she had 3 bags of cans, and the next day she found 7 more bags of cans and 44 bags of bottles. The question asks how many bags of cans she had altogether. The answer given in the image is 10 bags. By adding the number of bags of cans found on both days, we can conclude that Tiffany had a total of 10 bags of cans. This problem tests the ability to understand information from the question and perform simple addition.

\mytcbmanual{Dataloading for Interation 1}{

\textbf{Problem:} \hfill  \textcolor{gray}{YYYY-MM-DD 22:35:34,676 - log.logger\_utils - INFO}
\\[2pt]   
\textit{Tiffany was collecting cans for recycling. On Monday, she had 3 bags of cans. The next day, she found 7 more bags
of cans and 44 bags of bottles. How many bags of cans did she have altogether?}

\vspace{1em}

\textbf{Answer:} \hfill  \textcolor{gray}{YYYY-MM-DD 22:35:34,676 - log.logger\_utils - INFO}
\\[2pt]   
\textit{10}

}{python}

This example demonstrates the parameter settings, systematic problem-solving approach, and the beginning of the answer provided by the first agent when tackling a mathematical word problem. The focus is on the standardized process of problem-solving, emphasizing a complete workflow from understanding the question to checking the final answer.

\mytcbmanual{Training for Agent 1 in Interaction 1}{

\textbf{Prameter Log for Agent 1:} \hfill  \textcolor{gray}{YYYY-MM-DD 22:35:34,691 - log.logger\_utils - INFO}
\\[2pt]
temperature: 0.9193923355128871,\\
top\_p: 0.9027935433165666,\\
frequency\_penalty: 1.8091973063219,\\
presence\_penalty: -1.7443228132190376,\\
max\_tokens: 1919

\vspace{1em}

\textbf{prompt Log for Agent 1:} \hfill  \textcolor{gray}{YYYY-MM-DD 22:35:48,332 - log.logger\_utils - INFO}
\\[2pt]
To solve the questions found in the svamp dataset, follow these steps:\\
1. Carefully read the entire question to understand what is being asked.\\
2. Identify and extract all relevant numerical data and quantities mentioned within the question.\\
3. Determine which mathematical operations (addition, subtraction, multiplication, division) are necessary based on keywords or phrases that indicate relationships between numbers (e.g., "altogether," "more than," etc.).\\
4. Perform calculations step-by-step while keeping track of intermediate results if needed to avoid errors.\\
5. Double-check your final answer against the context of the problem to ensure it makes sense.\\
Use this approach systematically for each question from the SVAMP dataset you encounter!

\vspace{1em}

\textbf{Answer by Agent 1:} \hfill  \textcolor{gray}{YYYY-MM-DD 22:35:52,418 - log.logger\_utils - INFO}
\\[2pt]
Let's follow the steps to solve the problem:\\
1. Read the entire question: Tiffany collected cans for recycling, and we need to find out how many bags of cans she had altogether after her collection over two days.\\
2. Extract relevant numerical data:\\
On Monday, Tiffany had 3 bags of cans.\\
On Tuesday, she found 7 more bags of cans.\\
3. Determine necessary mathematical operations: We need to find the total number of bags of cans, which involves addition since we are combining quantities.\\
4. Perform calculations step-by-step:\\
Start with the number of bags she had on Monday: 3\\
Add the number of additional bags found on Tuesday:\\
\[
\text{Total Bags} = 3 + 7 = 10
\]\\
5. Double-check final answer against context: The question asks for how many bags of "cans" she has altogether, and we calculated that Tiffany has a total of 10 bags (only counting "cans").\\
Therefore, Tiffany has a total of 10 bags of cans altogether.

}{python}

This example provides a detailed demonstration of how the first agent systematically solves a mathematical word problem using a five-step approach (understanding the problem, extracting data, determining the operation, step-by-step calculation, and checking the answer), ultimately arriving at a clear and accurate solution.

\mytcbmanual{Training for Agent 2 in Interaction 1}{

\textbf{Prameter Log for Agent 2:} \hfill  \textcolor{gray}{YYYY-MM-DD 22:35:52,532 - log.logger\_utils - INFO}
\\[2pt]
temperature: 0.39241099147650554,\\
top\_p: 0.6602020470317261,\\
frequency\_penalty: 0.0007142275093004872,\\
presence\_penalty: -0.19136457572912707,\\
max\_tokens: 1413

\vspace{1em}

\textbf{prompt Log for Agent 2:} \hfill  \textcolor{gray}{YYYY-MM-DD 22:36:05,253 - log.logger\_utils - INFO}
\\[2pt]
Create a word problem that involves basic arithmetic operations based on the given scenario. The problem should include specific quantities and require calculations to determine totals, differences, or other values relevant to everyday activities, ensuring that it can engage learners in applying their math skills within relatable contexts like shopping, collecting items, or planning events.

\vspace{1em}

\textbf{Answer by Agent 2:} \hfill  \textcolor{gray}{YYYY-MM-DD 22:36:09,262 - log.logger\_utils - INFO}
\\[2pt]
To find out how many bags of cans Tiffany had altogether, we need to focus only on the bags of cans.\\
1. On Monday, Tiffany had 3 bags of cans.\\
2. The next day, she found 7 more bags of cans.\\
Now, we can calculate the total number of bags of cans:\\
Total bags of cans = Bags on Monday + Bags found on Tuesday\\
Total bags of cans = 3 + 7\\
Now, let's do the calculation:\\
Total bags of cans = 3 + 7 = 10\\
So, Tiffany had 10 bags of cans altogether.\\
(Note: The bags of bottles are not included in the total for bags of cans, as the problem specifically asks for the total number of bags of cans.)

}{python}

This example demonstrates how the second agent generates and solves a simple math word problem following a standardized process. It covers all key steps, including problem design, information extraction, mathematical operations, step-by-step calculation, and explanation of the result. The example also specifically clarifies the scope of the problem (counting only bags of cans, not bottles), which helps the agent understand the problem requirements and the solution approach.

\mytcbmanual{Training for Agent 3 in Interation 1}{

\textbf{Prameter Log for Agent 3:} \hfill  \textcolor{gray}{YYYY-MM-DD 22:36:09,369 - log.logger\_utils - INFO}
\\[2pt]
temperature: 0.5822851007143794,\\
top\_p: 0.9719972687061527,\\
frequency\_penalty: 0.04468780736250461,\\
presence\_penalty: 0.02409206816099846,\\
max\_tokens: 1397

\vspace{1em}

\textbf{prompt Log for Agent 3:} \hfill  \textcolor{gray}{YYYY-MM-DD 22:36:21,755 - log.logger\_utils - INFO}
\\[2pt]
You are given various scenarios involving numerical problems related to quantities, collections, and arithmetic operations. Your task is to answer questions based on the provided information for each scenario. Pay attention to details such as initial amounts, changes over time or circumstances (e.g., additions or subtractions), and specific requests regarding total counts. Use basic mathematical reasoning and calculations when determining your answers.

\vspace{1em}

\textbf{Answer by Agent 3:} \hfill  \textcolor{gray}{YYYY-MM-DD 22:36:24,662 - log.logger\_utils - INFO}
\\[2pt]
To find out how many bags of cans Tiffany had altogether, we need to focus only on the bags of cans.\\
On Monday, Tiffany had 3 bags of cans. The next day, she found 7 more bags of cans.\\
Now, we can calculate the total number of bags of cans:\\
Total bags of cans = Bags on Monday + Bags found on Tuesday\\
Total bags of cans = 3 + 7 = 10\\
Therefore, Tiffany had a total of 10 bags of cans altogether.

}{python}

Agent 3 focuses on generating simple and straightforward arithmetic word problems, usually centered on addition or subtraction, with scenarios drawn from everyday life for easy understanding. The problems and solutions are very concise, typically containing only key information without extra explanations or reasoning steps. This approach is suitable for foundational arithmetic instruction or quick practice, emphasizing the calculation itself rather than the problem-solving process.

\mytcbmanual{Training for Agent 4 in Interaction 1}{

\textbf{Prameter Log for Agent 4:} \hfill  \textcolor{gray}{YYYY-MM-DD 22:36:24,768 - log.logger\_utils - INFO}
\\[2pt]
temperature: 0.5298405602481979,\\
top\_p: 0.7372076916536928,\\
frequency\_penalty: 0.02248299928407254,\\
presence\_penalty: 0.049040988666425915,\\
max\_tokens: 1746

\vspace{1em}

\textbf{prompt Log for Agent 4:} \hfill  \textcolor{gray}{YYYY-MM-DD 22:36:37,855 - log.logger\_utils - INFO}
\\[2pt]
Create a word problem involving everyday scenarios that require basic arithmetic operations to solve. The scenario should include quantities related to common activities such as shopping, collecting items, or completing tasks over a period of time. Ensure the question asks for a total count based on additions and potential subtractions, utilizing numbers that are simple and relatable while encouraging mathematical reasoning in interpreting the context provided.

\vspace{1em}

\textbf{Answer by Agent 4:} \hfill  \textcolor{gray}{YYYY-MM-DD 22:36:41,333 - log.logger\_utils - INFO}
\\[2pt]
To find out how many bags of cans she had altogether, we only need to focus on the bags of cans.\\
1. Start with the number of bags of cans Tiffany had on Monday: 3 bags.\\
2. Add the number of bags of cans she found on Tuesday: 7 bags.\\
Now, calculate the total number of bags of cans:\\
\[
\begin{aligned}
3~\text{bags (Monday)} + 7~\text{bags (Tuesday)}
\\
= 10~\text{bags of cans}.
\end{aligned}
\]
Therefore, Tiffany had 10 bags of cans altogether.\\
(Note: The bags of bottles are not included in the total for cans.)

}{python}

Agent 4’s prompts emphasize problem generation that incorporates real-life scenarios, encouraging mathematical reasoning and the filtering of relevant information. Its answers are usually broken down into steps, highlighting the problem-solving approach and information extraction, but do not directly provide the final answer, focusing more on guiding the Agent to think independently. The overall style leans towards instructional guidance, helping the Agent understand the problem and develop independent problem-solving skills.

\mytcbmanual{Training for Agent 5 in Interation 1}{

\textbf{Prameter Log for Agent 5:} \hfill  \textcolor{gray}{YYYY-MM-DD 22:36:41,435 - log.logger\_utils - INFO}
\\[2pt]
temperature: 0.3603354498081722,\\
top\_p: 0.8427861799350946,\\
frequency\_penalty: 0.0572741656361528,\\
presence\_penalty: -0.124744427417488,\\
max\_tokens: 1736

\vspace{1em}

\textbf{prompt Log for Agent 5:} \hfill  \textcolor{gray}{YYYY-MM-DD 22:36:53,728 - log.logger\_utils - INFO}
\\[2pt]
Create a problem-solving question that involves basic arithmetic operations such as addition, subtraction, multiplication, or division within various contexts. The scenarios should include everyday situations involving quantities of objects or items to be calculated. Make sure the questions require an understanding of how to combine different amounts and solve for totals or differences based on the information provided in each scenario.

\vspace{1em}

\textbf{Answer by Agent 5:} \hfill  \textcolor{gray}{YYYY-MM-DD 22:37:00,748 - log.logger\_utils - INFO}
\\[2pt]
To solve the problem, we need to focus only on the bags of cans, as the question specifically asks for the total number of bags of cans Tiffany had.\\
1. On Monday, Tiffany had 3 bags of cans.\\
2. On Tuesday, she found 7 more bags of cans.\\
Now, we can add the number of bags of cans she had on Monday to the number she found on Tuesday:\\
\[
\begin{aligned}
\text{Total bags of cans} = \text{Bags on Monday} 
\\
+ \text{Bags found on Tuesday}.
\end{aligned}
\]
Substituting the values:\\
\[
\text{Total bags of cans} = 3 + 7
\]
Now, we perform the addition:\\
\[
\text{Total bags of cans} = 10
\]
Thus, Tiffany had a total of 10 bags of cans altogether.

}{python}

Agent 5’s prompts cover a variety of operations, including addition, subtraction, multiplication, and division, with a strong emphasis on information integration and analytical skills. Its solutions are very detailed, systematically presenting the problem-solving process step by step, including formulas, substitution, calculation, and conclusion, with formal and well-organized language.

\subsubsection{Evaluating Examples}

This section systematically evaluates the performance of the agent during a specific task iteration. The evaluation process adopts a structured scoring system, conducting detailed analysis and scoring across three core dimensions: logic, ingenuity, and completeness, ultimately resulting in a comprehensive total score. 

\mytcbmanual{Evaluating for Agent 1 in Interation 1}{

\textbf{Evalauation result:} \hfill  \textcolor{gray}{YYYY-MM-DD 22:37:10,792 - log.logger\_utils - INFO}

\vspace{1em}

\textbf{Logic (0 - 100):}\\
"The solution follows a clear logical structure. It begins with understanding the question and identifying key data points, followed by performing the correct mathematical operations to arrive at the answer. The steps are clearly articulated, ensuring that the reasoning is easy to follow."\\
(Score: 90)

\vspace{1em}

\textbf{Ingenuity (0 - 100):}\\
"While the solution is logical and methodical, it lacks any innovative or creative approaches to solving the problem. It simply adheres to standard arithmetic operations without introducing any unique methods or insights."\\
(Score: 70)

\vspace{1em}

\textbf{Completeness (0 - 100):}\\
"The answer comprehensively addresses all parts of the question, including reading the question, identifying numerical data, performing calculations, and verifying the final result. However, it could benefit from a brief summary reiterating the significance of reaching this total within the context."\\
(Score: 85)

\vspace{1em}

\textbf{Total Score (0 - 100):} (Score: 82)

}{python}

Specifically, the evaluator first reviews the agent’s solution process to determine whether it follows clear logical steps, including understanding the problem, extracting key data points, and performing correct mathematical calculations; based on this, a score is given for the logic dimension. Next, the evaluator examines whether the agent demonstrates unique or innovative approaches in problem-solving, or simply relies on conventional methods, and assigns a score for ingenuity accordingly. Finally, the evaluator checks whether the answer fully addresses all aspects of the problem, including data identification, calculation, and result verification, as well as whether the significance of the result is summarized, thus determining the completeness score. Each dimension is accompanied by a brief written assessment highlighting strengths and weaknesses, and a comprehensive total score is calculated based on an overall consideration. This evaluation method provides a holistic reflection of the agent’s problem-solving performance and offers concrete references for subsequent optimization and improvement.

\mytcbmanual{Evaluating for Agent 2 in Interaction 1}{

\textbf{Evalauation result:} \hfill  \textcolor{gray}{YYYY-MM-DD 22:37:19,822 - log.logger\_utils - INFO}

\vspace{1em}

\textbf{Logic (0 - 100):}\\
"The answer correctly identifies the number of bags of cans Tiffany had by logically breaking down the information provided. It accurately sums the bags from Monday and Tuesday without including irrelevant details about bottles."\\
(Score: 95)

\vspace{1em}

\textbf{Ingenuity (0 - 100):}\\
"The approach is straightforward and effective, but it does not demonstrate any creative or innovative problem-solving techniques. The solution follows a standard arithmetic process without additional insights."\\
(Score: 70)

\vspace{1em}

\textbf{Completeness (0 - 100):}\\
"The answer addresses the question fully by providing a clear calculation and stating that bags of bottles are not included in the total for bags of cans. However, it could have briefly summarized why only cans were considered to enhance clarity."\\
(Score: 85)

\vspace{1em}

\textbf{Total Score (0 - 100):} (Score: 83)

}{python}

\mytcbmanual{Evaluating for Agent 3 in Interation 1}{

\textbf{Evalauation result:} \hfill  \textcolor{gray}{YYYY-MM-DD 22:37:27,859 - log.logger\_utils - INFO}

\vspace{1em}

\textbf{Logic (0 - 100):}\\
"The solution correctly identifies the relevant information and follows a clear logical flow to arrive at the answer. The calculations are straightforward and accurate."\\
(Score: 90)

\vspace{1em}

\textbf{Ingenuity (0 - 100):}\\
"The solution employs basic arithmetic to solve a simple problem. While it is effective, there is little creativity or innovation in the approach."\\
(Score: 60)

\vspace{1em}

\textbf{Completeness (0 - 100):}\\
"The answer thoroughly addresses the problem, providing all necessary calculations and arriving at a final total. However, it could be enhanced by including a brief summary of the given information."\\
(Score: 85)

\vspace{1em}

\textbf{Total Score (0 - 100):} (Score: 78)

}{python}

\mytcbmanual{Evaluating for Agent 4 in Interation 1}{

\textbf{Evalauation result:} \hfill  \textcolor{gray}{YYYY-MM-DD 22:37:37,127 - log.logger\_utils - INFO}

\vspace{1em}

\textbf{Logic (0 - 100):}\\
"The reasoning presented in the answer is clear and follows a logical sequence. It identifies the relevant information (the number of bags of cans) and correctly excludes irrelevant information (the bags of bottles). The calculation is straightforward and accurate."\\
(Score: 90)

\vspace{1em}

\textbf{Ingenuity (0 - 100):}\\
"The answer demonstrates a basic understanding of arithmetic and problem-solving. However, it lacks creative or innovative approaches to the problem, as it simply applies basic addition without any additional insights or unique methods."\\
(Score: 50)

\vspace{1em}

\textbf{Completeness (0 - 100):}\\
"The answer provides a complete solution to the problem by addressing all necessary steps to arrive at the final count of bags of cans. It states clearly what is included in the count and what is not, making it easy to follow."\\
(Score: 90)

\vspace{1em}

\textbf{Total Score (0 - 100):} (Score: 76)

}{python}

\mytcbmanual{Evaluating for Agent 5 in Interation 1}{

\textbf{Evalauation result:} \hfill  \textcolor{gray}{YYYY-MM-DD 22:37:37,229 - log.logger\_utils - INFO}

\vspace{1em}

\textbf{Logic (0 - 100):}\\
"The solution follows a clear and logical sequence of steps to arrive at the answer. It correctly identifies the relevant information, uses appropriate arithmetic, and clearly states the final result."\\
(Score: 95)

\vspace{1em}

\textbf{Ingenuity (0 - 100):}\\
"The approach is straightforward and relies on basic arithmetic, which does not showcase high levels of creativity or innovative problem-solving. However, it effectively solves the problem without unnecessary complexity."\\
(Score: 70)

\vspace{1em}

\textbf{Completeness (0 - 100):}\\
"The answer provides a complete solution by including all necessary calculations and explanations. It explicitly states the initial quantities, performs the addition, and presents the final answer clearly."\\
(Score: 90)

\vspace{1em}

\textbf{Total Score (0 - 100):} (Score: 85)

}{python}

\subsubsection{Optimizing Examples}

This section illustrates the process of parameter updating during a particular iteration. Specifically, the system first ranks the agents participating in the current round of tasks and selects the top three performers. Next, the system updates the parameters of a certain agent (usually one that performed well or needs improvement). The parameter update involves adjusting key hyperparameters such as temperature, top\_p, frequency\_penalty, and presence\_penalty, all of which directly affect the diversity, creativity, and accuracy of the agent’s generated responses. The system records the specific values of these parameters before and after the update, and provides a task prompt to clarify the type of problem the agent needs to solve and any important considerations.

\mytcbmanual{Ranking for Iteration 1}{

\textbf{Ranking:} \hfill  \textcolor{gray}{YYYY-MM-DD 22:37:54,082 - log.logger\_utils - INFO}

\vspace{1em}

Top three results from message processing LLM:\\
"First Place": "2",\\
"Second Place": "5",\\
"Third Place": "1"

}{python}

\mytcbmanual{Updating for Agent in Iteration 1}{

\textbf{Before Updating:} \hfill  \textcolor{gray}{YYYY-MM-DD 22:37:54,082 - log.logger\_utils - INFO}

Parameter:\\
temperature=0.9193923355128871,\\
top\_p=0.9027935433165666,\\
frequency\_penalty=1.8091973063219,\\
presence\_penalty=-1.7443228132190376

prompt:\\
You are given various scenarios involving numerical problems related to quantities, collections, and arithmetic operations. Your task is to answer questions based on the provided information for each scenario. Pay attention to details such as initial amounts, changes over time or circumstances (e.g., additions or subtractions), and specific requests regarding total counts. Use basic mathematical reasoning and calculations when determining your answers.

\vspace{1em}

\textbf{After Updating:} \hfill  \textcolor{gray}{YYYY-MM-DD 22:38:02,220 - log.logger\_utils - INFO}

Parameter:\\
temperature=0.41004817918398145,\\
top\_p=0.9247874518938444,\\
frequency\_penalty=-0.0035787094690344773,\\
presence\_penalty=0.025675460113628396

prompt:\\
You are presented with various scenarios that involve numerical problems related to quantities and arithmetic operations. Your task is to answer questions based on the provided information for each scenario. Start by carefully reading each question to understand what is being asked. Identify all relevant numerical data and quantities mentioned. Determine which mathematical operations (addition, subtraction, multiplication, division) are needed based on keywords or phrases that indicate relationships between numbers. Perform calculations step-by-step, keeping track of intermediate results if necessary, and double-check your final answer against the context of the problem to ensure it makes sense. Pay attention to details such as initial amounts, changes over time, and specific requests regarding total counts, using basic mathematical reasoning and calculations to determine your answers.

}{python}

This entire process embodies an automated “evaluation–selection–fine-tuning–re-evaluation” optimization loop, aiming to continuously improve the agent’s performance on specific types of problems through iterative trial and adjustment.

\section{Responsible Use Statement}

\noindent\textbf{Reproducibility and transparency.}
We are committed to the full reproducibility of this work. The proposed Grey Wolf Optimizer (GWO)-based multi-agent collaboration framework, along with its core algorithms and iterative optimization procedure, is fully described with pseudocode in the Appendix to ensure that researchers can directly reproduce and extend our study.
Our experimental setup is detailed in Section ~\ref{Experiments}. We evaluate the framework on both mathematical reasoning and hybrid reasoning benchmarks. These publicly available datasets cover arithmetic, algebra, geometry, logic, and interdisciplinary reasoning, thereby ensuring diverse experimental validation.
For model selection, we employ several mainstream large language models to verify the robustness of our method across different scales and architectures.
To guarantee transparency, the algorithm pseudocode (including the standard GWO and the agent-based iterative optimization process) is provided in Appendix A, while experimental results, ablation studies, and additional performance comparisons are presented in Appendix A.4 and A.5. All experiments are implemented in a Python environment and executed on a multi-GPU system to ensure efficiency in both inference and training. We will release the complete source code and configuration files upon publication, enabling other researchers to directly verify and extend our results.

\noindent\textbf{THE USE OF LARGE LANGUAGE MODELS (LLMS)}
The authors utilized OpenAI’s GPT-5 to improve the grammar, clarity, and conciseness of the text. All scientific contributions, methodology design, experiments, and analyses are the original work of the authors, who take full responsibility for the paper’s content.

\end{document}